\documentclass[conference]{IEEEtran}
\IEEEoverridecommandlockouts
\usepackage{cite}
\usepackage[pdftex]{graphicx}

\usepackage{amsmath}
\usepackage{amssymb}
\usepackage[ruled]{algorithm2e}
\usepackage{floatrow}
\usepackage{multicol}
\usepackage{multirow}
\usepackage{dsfont}
\usepackage{booktabs}
\usepackage{bm}
\usepackage{wrapfig}
\usepackage{cleveref}
\let\labelindent\relax
\usepackage{enumitem}
\usepackage{mathtools}
\usepackage{subcaption}
\usepackage{tikz}
\usepackage{adjustbox}
\usepackage{floatflt}
\usetikzlibrary{calc}
\usepackage{wrapfig,lipsum,booktabs}
\usepackage[numbers]{natbib}
\usetikzlibrary{arrows}
\renewcommand\vec[1]{\mathbf{#1}}

\begin{document}
\tikzstyle{decision} = [diamond, draw, fill=blue!20, 
    text width=6em, text badly centered, node distance=3cm, inner sep=0pt]
\tikzstyle{block} = [rectangle, draw, fill=blue!20, 
    text width=6em, text centered, rounded corners, minimum height=3.2em]
\tikzstyle{line} = [draw, -latex']
\tikzstyle{cloud} = [draw, ellipse,fill=red!20, node distance=3cm,
    minimum height=2em]
% paper title
\title{Probabilistic Trajectory Prediction with Structural Constraints}

% You will get a Paper-ID when submitting a pdf file to the conference system
\author{Weiming Zhi$^{1,*}$, Lionel Ott$^{2}$, Fabio Ramos$^{1,3}$
\thanks{Correspondence to: W. Zhi, {\tt\small weiming.zhi@sydney.edu.au}.}%
\thanks{$^{1}$ School of Computer Science, the University of Sydney, Australia}%
\thanks{$^{2}$ Autonomous Systems Lab, ETH Zurich, Zurich, Switzerland}%
\thanks{$^{3}$ NVIDIA, USA}
}

\maketitle

\begin{abstract}
%On the other hand, trajectory optimisation allows for the explicit incorporation of constraints with respect to environment occupancy, but requires a well-defined cost function which typically does not incorporate information from observations. We propose a framework that integrates motion trajectory learning and optimisation to explicitly incorporate environment structure for motion prediction. We empirically demonstrate on real and simulated datasets the ability of our framework to enforce constraints, and produce high quality predictions. 
This work addresses the problem of predicting the motion trajectories of dynamic objects in the environment. Recent advances in predicting motion patterns often rely on machine learning techniques to extrapolate motion patterns from observed trajectories, with no mechanism to directly incorporate known rules. We propose a novel framework, which combines probabilistic learning and constrained trajectory optimisation. The learning component of our framework provides a distribution over future motion trajectories conditioned on observed past coordinates. This distribution is then used as a prior to a constrained optimisation problem which enforces chance constraints on the trajectory distribution. This results in constraint-compliant trajectory distributions which closely resemble the prior. In particular, we focus our investigation on collision constraints, such that extrapolated future trajectory distributions conform to the environment structure. We empirically demonstrate on real-world and simulated datasets the ability of our framework to learn complex probabilistic motion trajectories for motion data, while directly enforcing constraints to improve generalisability, producing more robust and higher quality trajectory distributions. %\fabio{You need more about the novelty of the method, eg, posterior regularisation for trajectory prediction, and less on the motivation. }
\end{abstract}
\maketitle
\section{Introduction}
Autonomous robots, such as service robots and self-driving vehicles, are often required to coexist in environments with other dynamic objects. For robots to safely navigate and plan in an anticipatory manner in dynamic environments, accurate motion predictions for other nearby objects are needed. Many recent developments \cite{Alahi2016SocialLH,Gupta2018SocialGS,wzhiKTM} in motion prediction have relied on learning based approaches, whereby future positions, or probabilities over future positions, are learned from motion trajectories of previously observed dynamic objects. Learning-based methods to motion prediction typically extract patterns from the training data. These patterns are then used to extrapolate future positions based on new observations without explicitly considering the feasibility of the prediction. 

However, there are often rules for motion trajectories that are known {\em a priori}, but are not explicitly enforced during extrapolation by learning-based models. In particular, we address the problem of incorporating constraints that arise from the environment's structure, such as obstacles, into motion predictions. In most environments, motion trajectories are not highly unstructured, and depend closely on the environment occupancy structure. 

To address the shortcomings of purely learning-based methods, we view motion prediction as a combined probabilistic trajectory learning and constrained trajectory optimisation problem. Constrained optimisation methods allow for constraints to be imposed directly onto the distribution of trajectories without incorporating the collected data. We propose a novel framework that incorporates structural constraints into probabilistic motion prediction problems. Our framework leverages the ability of learning based models to extract patterns from data, as well as the ability of constrained trajectory optimisation approaches to explicitly specify constraints, giving better generalisability and higher quality extrapolations. Specifically, our contributions consist of: (1) a method to learn a distribution over trajectories from observations that is amenable to being optimised; (2) an approach to enforce constraints on distributions of trajectories, particularly collision chance constraints when provided an occupancy map.

\begin{figure}[t]
    \centering
    \includegraphics[width=0.75\textwidth]{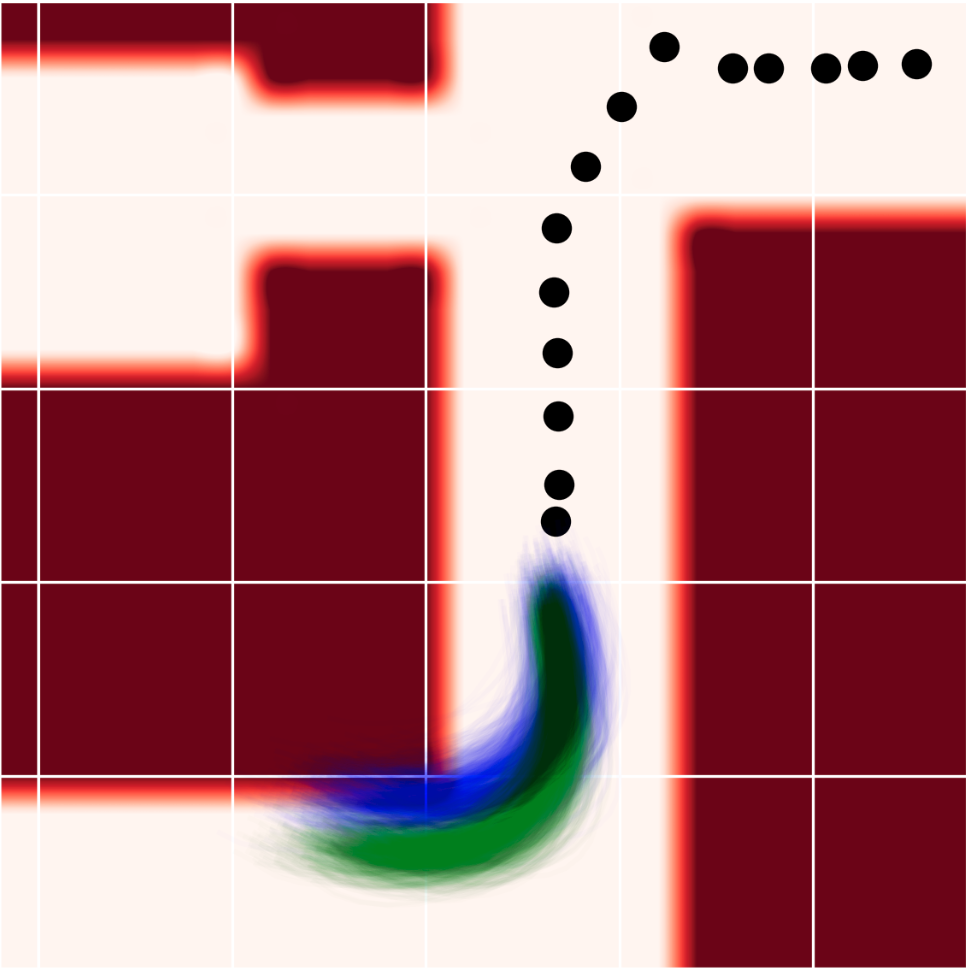}
    \caption{A simple example of enforcing predictions to be consistent with environment: Provided an observed trajectory segment (black), a purely learning method may predict collision-prone distributions of trajectories (blue). Our framework formulates an optimisation problem with chance constraints arising from environmental structure, as given by an occupancy map. We solve to obtain predictions which are compliant with the environment, while remaining similar to the learned future.}
    \label{Fig1}
\end{figure}

\section{Related Work}\label{RelatedWork}
\textbf{Motion Trajectory Prediction.}
Motion prediction is a problem central to robot autonomy. The most simple methods of motion prediction are physics-based models, such as constant velocity/acceleration models \cite{Schubert2008ComparisonAE}. More complex physics-based models often fuse multiple dynamic models, or select a best fit model from a set \cite{Kooij2018ContextBasedPP}. Recent developments in machine learning have led to learning-based methods that learn motion patterns \cite{Alahi2016SocialLH,Zhi2019SpatiotemporalLO} from a dataset of observed trajectories. In particular, long-short term memory (LSTM) and generative adversarial network (GAN)-based models \cite{Alahi2016SocialLH,Zhang2019SRLSTMSR,Amirian2019SocialWL,Gupta2018SocialGS} have gained popularity. Methods of learning entire trajectories, similar to our learning component, but with relaxed dependencies, is outlined in \cite{wzhiKTM,zhiOct}. Further attempts to incorporate static obstacles in learning attempt to do so as a part of the learning pipeline \cite{Sadeghian_2019_CVPR} and attempts to extrapolate behaviour. Approaches of this kind cannot guarantee that valid solutions satisfy specified chance constraints, and there have not been approaches that directly incorporate occupancy maps, which are typically built by robots from depth sensor data.\\
%Attempts to encode environmental factors through learning include, various flow map methods \cite{McCalman2013MultimodalEW,Kucner2017EnablingFA,Senanayake2018DirectionalGM,Mellado2019GoWT}, which try to predict the typical direction agents move in. These methods are able to implicitly learn environment factors from the data, such as following existing paths. These are not able to explicitly factor optimise for collision avoidance, and often result in predicting motion which collide into the environment.\\
\textbf{Trajectory Optimisation.}
Trajectory optimisation has been successfully applied in motion planning, by directly expressing obstacles as constraints to solve for collision-free paths between start and goal points. Popular methods include \emph{Trajopt} \cite{Schulman2013FindingLO}, \emph{CHOMP} \cite{Zucker2013CHOMPCH}, \emph{STOMP} \cite{Kalakrishnan2011STOMPST}, and \emph{GPMP2} \cite{Dong2016MotionPA}. Trajopt utilises sequential quadratic programming to find locally optimal trajectories compliant with constraints. CHOMP exploits obstacle gradients for efficient optimisation. Further work on gradient-based trajectory optimisation for planning can be found in \cite{Marinho2016FunctionalGM,Francis2019FastSF}. Although trajectory optimisation is a popular choice to obtain collision-free trajectories for motion-planning, there has yet to be extensive work in utilising trajectory optimisation for motion prediction. The optimisation required in this work is also different in that we optimise to obtain distributions of trajectories, rather than a single trajectory as required in motion planning.\\
\textbf{Combined Learning and Optimisation.} 
There has been previous explorations of combining learning and trajectory optimisation. \cite{Toussaint1} introduce a framework which combines a known analytical cost function with black box functions in reinforcement learning for manipulation. \cite{Rana2017TowardsRS} introduces CLAMP, a framework that combines learning from demonstration with optimisation based motion planning. Theoretical aspects of using learning in conjunction with optimisation, in a Bayesian framework known as posterior regularisation has also been investigated in \cite{PReg}. All of these methods aim to add more structure to be encoded in learning problems. Similarly our proposed framework allows for encoding of structure, but specifically for the trajectory prediction problem, where we are required to do learning and optimisation on distributions of trajectories with respect to environment occupancy.

\section{Continuous-time Motion Trajectories}\label{continuousTsec}
Our framework operates over continuous motion trajectories, capable of being queried at arbitrary resolution, without forward simulating. In this section, we briefly introduce the continuous-time motion trajectory representation used in this work. We restrict ourselves to the 2-D case. 

Similar to methods described in \cite{Francis2019FastSF,Marinho2016FunctionalGM,wzhiKTM}, we represent motion trajectories as functions $\bm{\xi}: [0,T]\rightarrow \mathbb{R}^{2}$. $T$ represents the prediction time horizon. To we construct our trajectories as dot products between weight vectors and \emph{reproducing kernels}. We limit our discussion to squared exponential radial basis function (RBF) kernels, though methods in this study generalise to various kernels with minimal modifications. Using RBFs result in smooth trajectories, and have shown good empirical convergence properties for functional trajectory optimisation \cite{Marinho2016FunctionalGM}. A trajectory, $\bm{\xi}(t)$, can be expressed as:
\begin{align}
    \vec \xi(t)\! = 
    \begin{bmatrix}
        \vec w^{x^\top} \bm \phi(t) \\
        \vec w^{y^\top} \bm \phi(t)
    \end{bmatrix},
    &&
    \vec \phi(t)\! =\! \left[
        \phi(t, t'_{1}), \dots, \phi(t, t'_M)
    \right]^\top,
    \label{trajDef}
\end{align}
where the basis functions are given as, $\phi(t, t') \!=\!\exp(-\gamma ||t - t'||^2)$. The coordinates of a dynamic object at time, $t\in[0,T]$, is given as $x(t)$ and $y(t)$, $\vec{w}^{x}$, $\vec{w}^{y}\in \mathbb{R}^M$ are weight parameters, $\bm{\phi}(t)\in \mathbb{R}^M$, and $\bm{t'}=[t'_1,t'_2,\ldots,t'_{M}]^{\top}$ are $M$ pre-define time points at which RBF kernels are centered. $\gamma$ is a length-scale hyper-parameter which determines the impact a time point neighbouring times on the trajectory. 

Motion data typically comes in the form of discrete sequences of timestamped coordinates $\{t_i, x_i,y_i\}, i = 0, \dots, N$. To construct a best-fit continuous trajectory of discrete sequences from $N$ timestamped coordinates, we solve the ridge regression problem:
 \begin{align}
     \min\limits_{\vec{w}^{x},\vec{w}^y}\sum\limits_{n=1}^{N}\big\{(x_n-{\vec{w}^{x}}^{\top}\bm{\phi}(t_n))^{2}&+(y_n-{\vec{w}^{y}}^{\top}\bm{\phi}(t_n))^{2}\nonumber\\
     &+\ldots\lambda(||\vec{w}^x||^{2}+||\vec{w}^y||^{2})\big\},\label{ridgeeqn}
 \end{align}
%\begin{equation}
%{\min\limits_{\bm{\omega}^{x},\bm{\omega}^y}\bigg\{ %\sum\limits_{n=1}^{N}\bigg(\begin{bmatrix}x_n\\y_n\end{bmatrix}-\begin{bmatrix}{\bm{\omega}^{x}}^{\top}\bm%{\phi}(t_n)\\{\bm{\omega}^{y}}^{\top}\bm{\phi}(t_n)\end{bmatrix} \bigg)^2 + %\lambda\begin{bmatrix}||\bm{\omega}^x||^{2}\\||\bm{\omega}^y||^{2}\end{bmatrix}\bigg\}},\label{ridgeeqn}
%\end{equation}
where $\lambda$ is a regularisation hyper-parameter, with the set of $M$ uniform times $\bm{t}'$ selected \emph{a priori}, as centers where the kernels are fixed. Evaluating the closed form ridge regression solution results in weight vectors $\vec{w}^{x}$ and $\vec{w}^{x}$ which forms the trajectory $\bm{\xi}(t)=[{\vec{w}^x}^{\top}\bm{\phi}(t),{\vec{w}^y}^{\top}\bm{\phi}(t)]$. We denote the stacked vector of $\vec{w}^{x}$ and $\vec{w}^{y}$ as $\vec{w}$. As future motion trajectories are inherently uncertain, the motion prediction problem often requires predicting a distribution over future trajectories a dynamic object may follow. We denote distributions of trajectories as $\bm{\Xi}$, with the vector of probabilistic weight parameters, $\bm{\omega}$, and the parameters of the distribution over $\bm{\omega}$ is denoted as $\Psi$. $\bm{\Xi}(t)$ gives the distribution of coordinate position at time $t$. Every sample function of $\bm{\Xi}$ is a trajectory $\bm{\xi}$, and samples of vector of random variables $\bm{\omega}$ give $\vec{w}$. Denoting the parameter distributions associated with $x$ and $y$ as $\bm{\omega}_{x}$ and $\bm{\omega}_{y}$, distributions of trajectories can be constructed as $\bm{\Xi}=[{\bm{\omega}^x}^{\top}\bm{\phi}(t),{\bm{\omega}^y}^{\top}\bm{\phi}(t)]$. In practice, to obtain $\bm\xi$ we can also sample $\vec{w}$ from the distribution $\bm{\omega}$, and take dot products with $\bm{\phi}$.

%$\bm{\Xi}=[{\bm{\omega}^x}^{\top}\bm{\phi}(t),{\bm{\omega}^y}^{\top}\bm{\phi}(t)]$, where $\bm{\Xi}$ is a stochastic process and $\bm{\Xi}(t)$ gives the distribution of coordinate position at time $t$, while every sample function of $\bm{\Xi}$ is a trajectory $\bm{\xi}$. Distributions over future trajectories can be modelled by assuming distributions over the weight parameters $\vec{w}$, with $\bm{\omega}$ denoting the vector of random variables, and distribution parameters $\Psi$. Realisations of $\bm{\omega}$ give $\vec{w}$. 
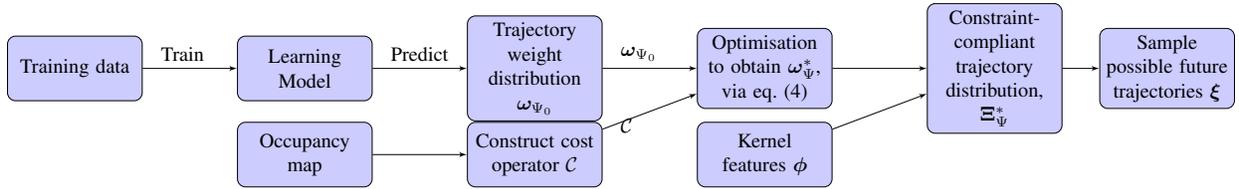
\begin{figure*}[t]
\begin{adjustbox}{width=0.9\textwidth,center} 
\begin{tikzpicture}[node distance = 1.6cm]
    % Place nodes
    \node [block] (Data){Training data};
    \node [block, right of=Data, node distance=4cm] (Pred) { Learning Model};
    \node [block, right of=Pred, node distance=4cm] (Params) { Trajectory weight distribution $\bm{\omega}_{\Psi_{0}}$};
    \node [block, below of=Pred, node distance=1.5cm] (Occ){ Occupancy map};
    \node [block, below of=Params, node distance=1.5cm] (Obs_cost) { Construct cost operator $\mathcal{C}$};
    \node [block, right of=Params, node distance=4cm] (Optimise){ Optimisation to obtain $\bm{\omega}_{\Psi}^{*}$, via \cref{formulationWhole}};
    \node [block, right of=Optimise, node distance=4cm] (Generate){ Constraint-compliant trajectory distribution, $\bm{\Xi}_{\Psi}^{*}$};
    \node [block, right of=Generate, node distance=3cm] (Sample){ Sample possible future trajectories $\bm{\xi}$};
    \node [block, below of=Optimise, node distance=1.5cm] (featuremaps){Kernel features $\bm{\phi}$};
    
    %\draw[red,thick,dotted] ($(Pred.north west)+(-0.2,0.2)$) node [text width=4cm,above]{Learning Component} rectangle ($(Params.south east)+(0.2,-0.2)$);
    
    % Draw edges
    \path [line] (Data) -- node [text width=1cm,midway,above]{Train}(Pred);
    \path [line] (Occ) -- (Obs_cost);
    \path [line] (Pred) -- node [text width=1cm,midway,above]{Predict}(Params);
    \path [line] (Obs_cost) -- node [text width=1cm,midway,below]{$\mathcal{C}$}(Optimise);
    \path [line] (Params) -- node [text width=1cm,midway,above]{$\bm{\omega}_{\Psi_{0}}$}(Optimise);
    \path [line] (Optimise) -- (Generate);
    \path [line] (featuremaps) -- (Generate);
    \path [line] (Generate) -- (Sample);

\end{tikzpicture}
\end{adjustbox}
\caption{Overview of our proposed framework. We combine learning and optimisation to generalise data and enforce constraints.}\label{Overview}
\end{figure*}

\section{Motion Prediction with Environmental Constraints}
\subsection{Problem Overview}
\begin{figure}[t]
  \begin{center}
    \includegraphics[width=0.35\textwidth]{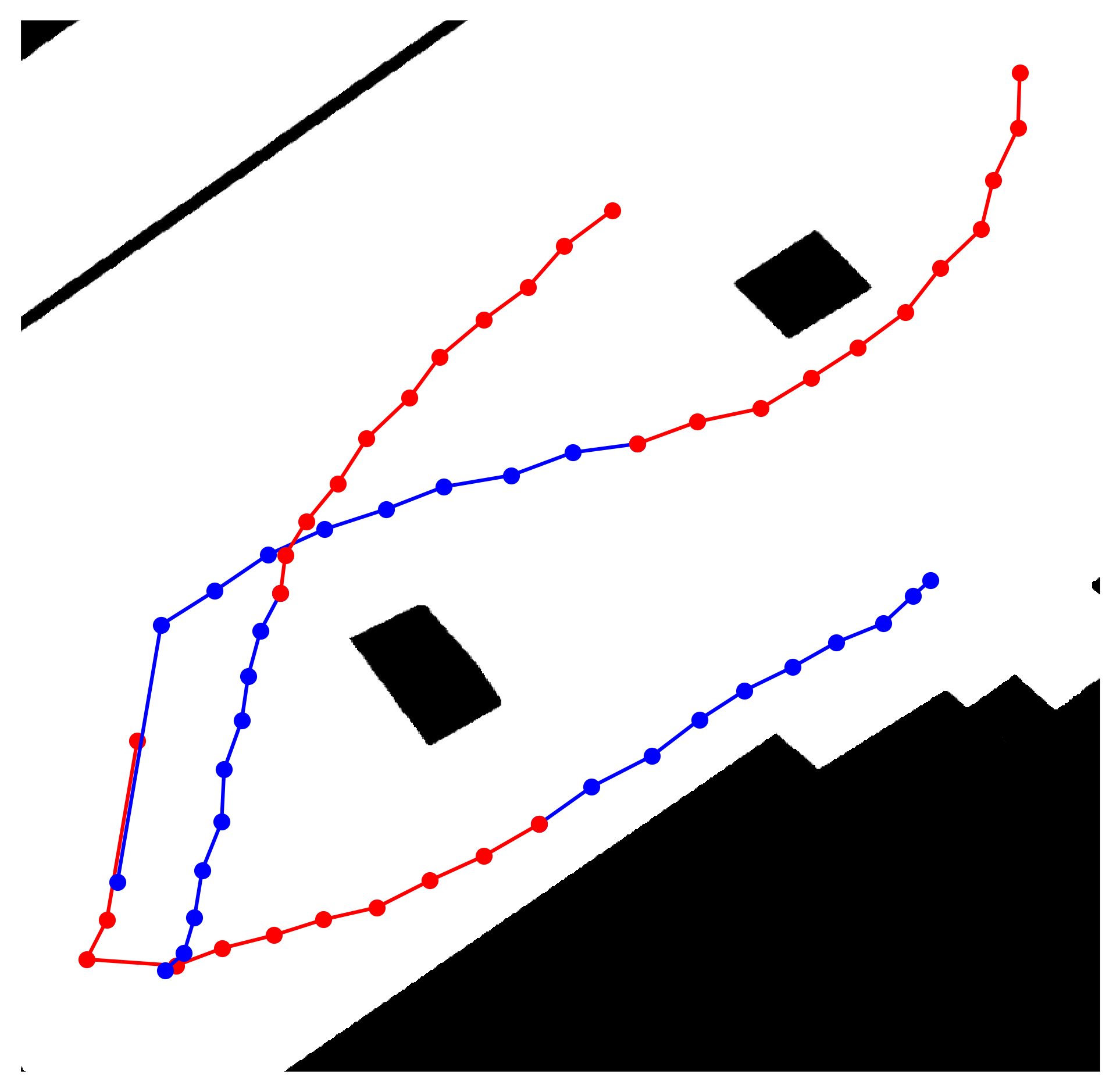}
    \includegraphics[width=0.35\textwidth]{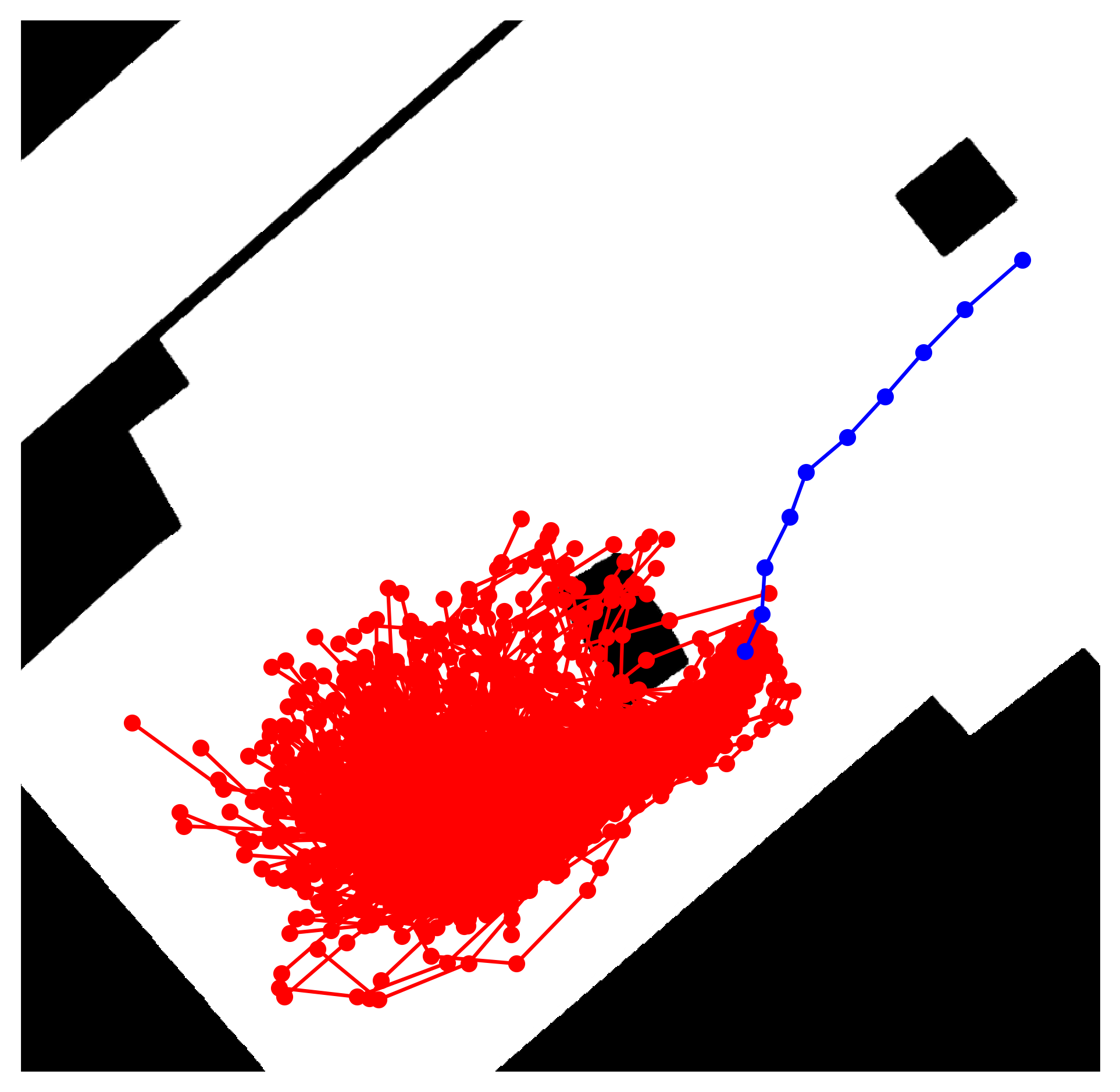}
  \end{center}
  \caption{(Left) Examples from real-world dataset \cite{thorDataset2019}. Historical trajectories in blue, and future in red. (Right) Provided a new unseen trajectory (blue), extrapolate futures (red). Learned future trajectories may collide with obstacles. Our framework tackles this by directly enforcing constraints on trajectory distributions, such that the trajectories are collision free. }\label{datasetProb}
\end{figure}
Our goal is to reason about the possible future trajectories of dynamic objects, based on the immediate past trajectories of the objects, along with occupancy information from the environment. In this work, we limit our discussion to trajectories in 2-D, which usually arise from the tracking of humans and vehicles in video data. We are assumed to have a dataset containing coordinates of an object up to a time, along with its coordinates thereafter, and an occupancy map of the environment. Examples from a real-world dataset are shown in \cref{datasetProb}, as well as a prediction given by a learning-based model. Given the observed historical coordinates, we aim to extrapolate the distribution of trajectories representing likely future motion thereafter. We require the distribution of trajectories compliant with the environment structure indicated in the occupancy map, such that the probability of hypothesising infeasible trajectories to be constrained to a small value.

\subsection{Framework Overview}
% \path [line] (Obs_cost) -- node [text width=2cm,left,below]{For constraint \\${\mathcal{C}\bm{\Xi}_{\Psi})-\epsilon\leq 0}$}(Optimise);
    %\path [line] (Params) -- node [text width=3cm,left,above]{For objective \\ $\min_{\Psi}\mathrm{D}_{KL}(\bm{\omega}_{\Psi}\lVert\bm{\omega}_{\Psi_{0}})$}(Optimise);
%Our framework takes a combined trajectory learning and optimisation approach to provide more structure in motion prediction problems. Commonly used machine learning models are capable of emulating training data, but typically lack mechanisms to directly enforce adherence to constraints. On the other hand, trajectory optimisation methods attempt to solve an optimisation problem to obtain trajectories that minimise a predefined cost while satisfying some constraints, without incorporating any collected data. Trajectory optimisation typically cannot be used directly for motion prediction, as the motion patterns of moving agents are often difficult to capture accurately in a specified cost function. Therefore, we view learning and trajectory optimisation as complementary for motion prediction.
% For each matrix normal distribution, an equivalent multivariate normal is formulated as, $\bm{\omega}\sim\mathcal{N}(\mathrm{vec}(\mathcal{M}),\mathcal{V}\otimes\mathcal{U})$, where $\mathrm{vec}(\cdot)$ is the vectorise operator, and $\otimes$ is the Kronecker product

An overview of our framework is shown in \cref{Overview}. We learn from our training data to predict a vector of random variables of weights $\bm{\omega}_{\Psi_{0}}$ with distribution parameters $\Psi_{0}$, and produce a prior distribution of trajectories, $\bm{\Xi}_{0}$, as described in \cref{trajDef}. Our aim is to find an alternative random vector $\bm{\omega}_{\Psi}^{*}$, defined by optimised parameters $\Psi^{*}$, which satisfies enforced constraints, while remaining close to the learned distribution $\bm{\omega}_{\Psi_{0}}$. The optimisation objective uses $\bm{\omega}_{\Psi_{0}}$, while constraints are provided through the \textit{obstacle cost operator}, $\mathcal{C}$, which requires an occupancy map of the environment. The desired random vector $\bm{\omega}_{\Psi}^{*}$ defines a constraint-compliant distribution of trajectories, $\bm{\Xi}_{\Psi}^{*}$, provided pre-defined RBF features $\bm{\phi}$, by \cref{trajDef}. Individual constraint-compliant future trajectories, $\bm{\xi}$, can then be realised from $\bm{\Xi}_{\Psi}^{*}$. 

To find the constrain-compliant parameters $\Psi^{*}$, we formulate the optimisation problem in a similar manner to Bayesian posterior regularisation methods \cite{PReg}. The desired parameters $\Psi^{*}$ is the solution to the constrained optimisation problem:
\begin{align}
    \min_{\Psi}\quad\mathrm{D}_{KL}(\bm{\omega}_{\Psi}\lVert\bm{\omega}_{\Psi_{0}}) \\ \textrm{s.t.}\quad  \mathcal{C}(\bm{\Xi}_{\Psi})-\epsilon\leq 0 \label{formulationWhole}
\end{align}
where $\mathrm{D}_{KL}$ is the Kullback-Leibler (KL) divergence \cite{BishopBook}, and $\mathrm{D}_{KL}(\bm{\omega}_{\Psi}\lVert\bm{\omega}_{\Psi_{0}})$ gives us a definition of ``close-ness'' between our predicted $\bm{\omega}_{\Psi_{0}}$ and optimised $\bm{\omega}_{\Psi}$ distributions. $\mathcal{C}(\bm{\Xi}_{\Psi})$ is the obstacle cost operator, which maps distributions of trajectories $\bm{\Xi}_{\Psi}$ to a scalar cost value. The cost is interpreted as the time-averaged probability of generating a trajectory that collides with obstacles. We designate $\epsilon$ to be the allowed limit of the obstacle cost. Intuitively, our optimisation problem returns the closest distribution of trajectories, as defined by the KL divergence term, to that given by our predictive model, subject to the obstacle collision constraint. By separating the learning and optimisation, we disentangle the complexities of the learning model and the structural constraints. Our learned $\bm{\omega}_{\Psi_{0}}$ also provides us with a ``warm-start'' initial solution to our optimisation. In contrast to learning-based approaches, the constraints in our optimisation problem is enforced in valid solutions. The main challenges to solving this trajectory optimisation problem lies in how to obtain the vector of random variables $\bm{\omega}_{\Psi_{0}}$, which defines our prior trajectory distribution $\bm{\Xi}_{0}$, as well as how to derive the obstacle cost operator $\mathcal{C}$ required for optimisation. These are addressed in the following subsections.

\subsection{Learning Distributions of Trajectories}\label{Learndists}
In this subsection, we outline a method of learning a mapping from observed historical trajectories to prior distributions of future trajectories, parameterised by $\bm{\omega}_{\Psi_{0}}$. The resulting prior distributions of trajectories are to be used directly in the optimisation problem \cref{formulationWhole}. We model the distribution of trajectories, $\bm{\Xi}$, as mixtures of multi-output stochastic processes. To capture the correlation between x and y coordinates, the distribution over weight parameters are assumed to be mixtures of matrix normal distributions \cite{Dawid1981SomeMD}. A matrix normal distribution is the generalisation of the normal distribution to matrix-valued random variables, where each sample from a matrix normal distribution gives a matrix. We work with the matrix of random variables defined by vertically stacking the parameters of coordinate dimensions, denoted as $\mathcal{W}=[\bm{\omega}^{x},\bm{\omega}^{y}]\in\mathbb{R}^{M\times 2}$:
\begin{align}
    \mathcal{W}=
    \begin{bmatrix}
    {w}^{x}_{1} & {w}^{y}_{1}\\
    \smash{\vdots} & \smash{\vdots}\\
    {w}^{x}_{M} & {w}^{y}_{M}
    \end{bmatrix}, &&
     p(\mathcal{W})&=\sum_{r=1}^{R}\alpha_{r}\mathrm{MN}\Big(
    \mathcal{M}_{r}, \mathcal{U}_{r}, \mathcal{V}_{r}
    \Big).
%    \text{where } 
%    p\Big(\begin{bmatrix}
%    \bm{\omega}^{x}\\
%    \bm{\omega}^{y}
%    \end{bmatrix}\Big)=&\sum_{r=1}^{R}\alpha_{r}\mathcal{N}\Bigg(
%    \begin{bmatrix}
%    \bm{\mu}^{x}_{r}\\
%    \bm{\mu}^{y}_{r}
%    \end{bmatrix},
%    \begin{bmatrix}
%    \Sigma^{x}_{r} & \Sigma^{x,y}_{r}\\
%    \Sigma^{y,x}_{r} & \Sigma^{y}_{r}
%    \end{bmatrix}
%    \Bigg)\label{E7}.
\end{align}
We assume that there are $R$ mixture components, similar to a Gaussian mixture model \cite{BishopBook}, with each component being a matrix normal distribution, and associated weight components $\alpha_r$. Matrix normal distributions have location matrices $\mathcal{M}_{r}=[\bm{\mu}_{r}^{x},\bm{\mu}_{r}^{y}]\in\mathbb{R}^{M \times 2}$, and positive-definite scale matrices $\mathcal{U}_{r}\in\mathbb{R}^{M \times M}$, $\mathcal{V}_{r}\in\mathbb{R}^{2 \times 2}$ as parameters \cite{MLEMatN}. 

We utilise a neural network to learn a mapping from a fixed-length vectorised coordinate sequences, $\bm{\varphi}$, that encodes observed partial trajectories, to component weights $\bm{\alpha}$, along with location matrices $\mathcal{M}_{r}$ the lower triangular $\mathcal{V}_{r}^{\frac{1}{2}}$ and diagonal $\mathcal{U}_{r}^{\frac{1}{2}}$ matrices. $\mathcal{V}$ is positive-definite and will be constructed by taking the product of two lower triangular matrices. The dependence between different times on the trajectory are largely captured the RBF features in the time domain $\bm{\phi}(t)$, as described in \cref{continuousTsec}. Hence, it is sufficient for $\mathcal{U}$ to be a positive definite diagonal matrix. We obtain the input vector $\bm{\varphi}$ by taking a 10 time-step window of the latest observed coordinates, and vectorising to obtain a vector of 20 dimensions. The neural network used is relatively light-weight, consisting of 4 Relu-activated dense layers with number of neurons: $(15\times M \times R)\rightarrow (5\times M \times R)\rightarrow (5\times M \times R)\rightarrow (R\times [3M+3])$. In this work, our main goal is to outline a representation of future trajectories outputted by the network, such that the result can be directly optimised. Our method does not preclude alternative encoding schemes of conditioned history trajectories to be inputted into the network, such as features from LSTM networks \cite{LSTM} or trajectory features outlined in \cite{wzhiKTM}. 

Provided a dataset of history and future trajectories, given by $N$ pairs of timestamped coordinate sequences. After flattening the history trajectory sequence to obtain $\bm{\varphi}$, and applying \cref{ridgeeqn} on trajectory futures to obtain weight matrices, we have pairs denoted as $\{\bm{\varphi}_{n}, \mathcal{W}_n\}_{n=1}^{N}$. We learn a mapping from $\bm{\varphi}_n$ to parameters $\{\alpha_{r},\mathcal{M}_{r},\mathcal{V}_{r},\mathcal{U}_{r}\}_{r=1}^{R}$ with the fully-connected neural network, with the negative log-likelihood of the weight matrices distributions as the loss function:
\begin{align}
    \mathcal{L}
    &=-\sum_{n=1}^{N}\log \nonumber\\ &\dots\sum_{r=1}^{R}\alpha_{r}\frac{\exp\{-\frac{1}{2}\mathrm{tr}[{\mathcal{V}_{r}}^{-1}\!(\mathcal{W}_n-\mathcal{M}_{r})^{\top}\!\mathcal{U}_{r}(\mathcal{W}_n-\mathcal{M}_{r})]\}}{(2\pi)^{M}|\mathcal{V}_{r}||\mathcal{U}_{r}|^{\frac{M}{2}}}.\label{fullLL}
\end{align}
and by applying the activation functions:
\begin{align}
    \mathrm{diag}[(\mathcal{U}_{r})^{\frac{1}{2}}]&=\exp(z^{\mathrm{diag}(\mathcal{U}_{r})}), \\ \mathrm{LowerTrig}[(\mathcal{V}_{r})^{\frac{1}{2}}]&=z^{\mathrm{LowerTrig}(\mathcal{V}_{r})}, \\
    \mathrm{diag}[(\mathcal{V}_{r})^{\frac{1}{2}}]&=\exp(z^{\mathrm{diag}(\mathcal{V}_{r})}), \\
    \alpha_{r}&=\mathrm{softmax}(z^{\alpha_r}),
\end{align}
where $z^{\alpha_r},z^{\mathrm{LowerTrig}(\mathcal{V}_{r})},z^{\mathrm{diag}(\mathcal{V}_{r})},z^{\mathrm{diag}(\mathcal{U}_{r})}$ are neural network outputs, and $\mathrm{diag}(\cdot)$, $\mathrm{LowerTrig}(\cdot)$ indicate the matrix diagonal, and strictly lower triangular matrix elements respectively. The $\exp$ activation function guarantees the validity of the scale matrices, by enforcing $z^{\mathrm{diag}(\mathcal{V}_{r})}>0, z^{\mathrm{diag}(\mathcal{U}_{r})}> 0$. We also enforce $\sum_{r=1}^{R}\alpha_{r}=1$ using the $\mathrm{softmax}$ function. After we train the neural network, provided a vectorised historical trajectory, $\bm{\varphi}$, the neural network gives the corresponding random vector of weights of our prior trajectory distribution, $\bm{\omega}_{\Psi_{0}}=\mathrm{vec}(\mathcal{W})$, where $\mathrm{vec}(\cdot)$ is the vectorise operator. The prior distributions of trajectories is then the dot product of the trajectory parameters and the RBF features, $\bm{\phi}(t)$, as described in \cref{trajDef}.

\subsection{Constraints on Distributions of Trajectories}\label{CostoperatorSubs}
After obtaining the weights, $\bm{\omega}_{\Psi_{0}}$, of a prior distribution of trajectories via learning, our main challenge is to design a cost operator, $\mathcal{C}(\bm{\Xi})$, to encode our desired constraints. Specifically we wish to design $\mathcal{C}(\bm{\Xi})$ to account for collisions of distributions of trajectories, provided an occupancy map. $\mathcal{C}(\bm\Xi)$ maps a distribution of smooth trajectories, $\bm{\Xi}$, to a real-valued scalar cost, which quantifies how collision-prone a given $\bm\Xi$ is. $\mathcal{C}$ differs from the cost operator in trajectory optimisation motion planning problems, such as those in \cite{Zucker2013CHOMPCH,Marinho2016FunctionalGM,Francis2019FastSF}, in that the input is a distribution over trajectories rather than a single trajectory. Our optimisation can not only optimise over trajectory mean, but also over trajectory variances.

\begin{figure}[t]
\floatbox[{\capbeside\thisfloatsetup{capbesideposition={right,top},capbesidewidth=0.3\textwidth}}]{figure}
{\caption{The abscissae (black) for estimating the distribution in world space with the Gauss-Hermite quadrature scheme}\label{fig:absc}}
{\includegraphics[width=0.52\textwidth]{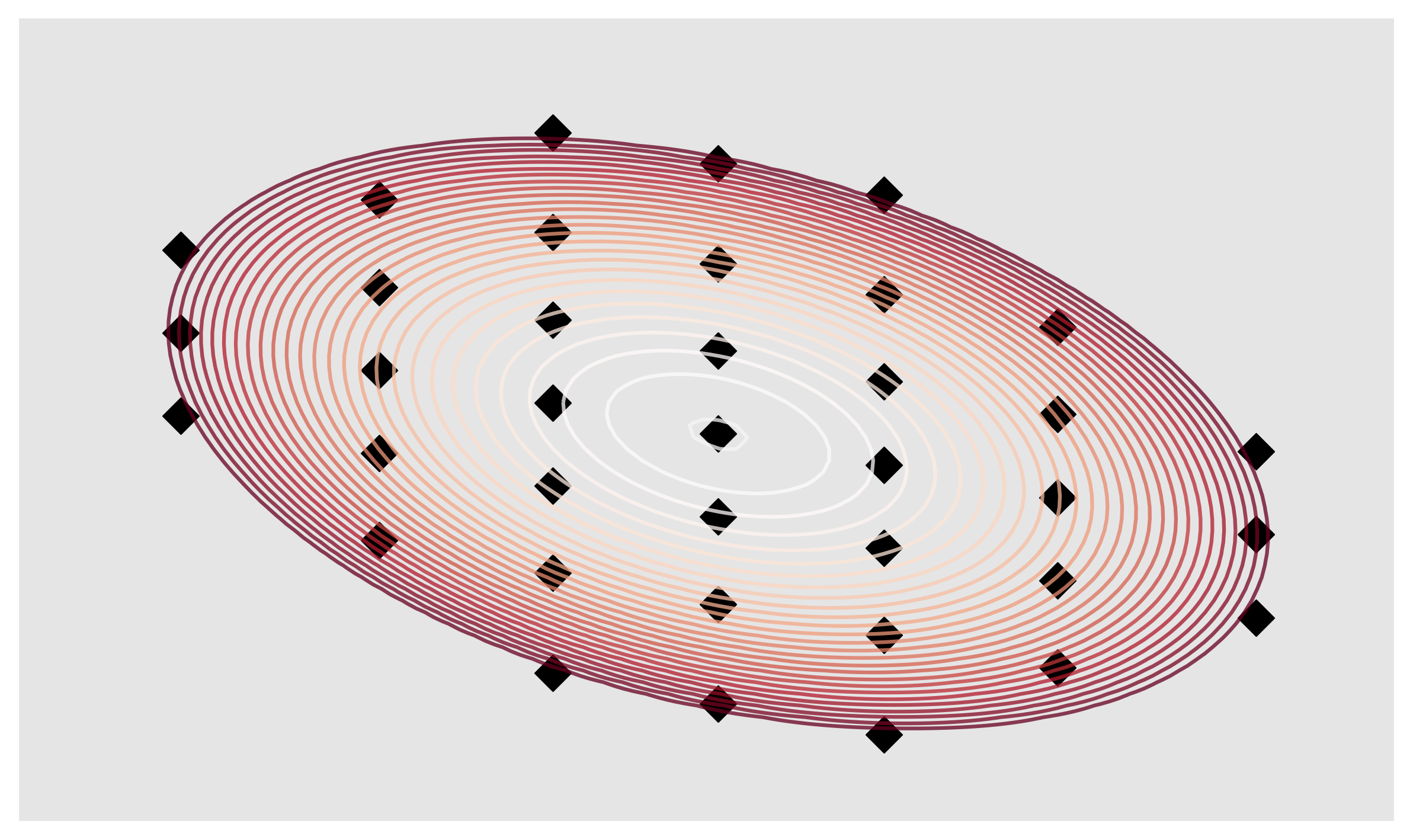}}
\end{figure}

Like trajectory optimisation methods \cite{Zucker2013CHOMPCH}, we assume the cost operator takes the form of average costs over prediction horizon $T$, where the cost at a given point in time, $\mathcal{C}(\bm{\Xi}_{\Psi}(t))$, is the probability of a collision at a given time. $\bm{\Xi}_{\Psi}$ contains weight variables with a distribution parameterised by $\Psi$. We can then interpret $\mathcal{C}(\bm{\Xi}_{\Psi})$ as the average probability of collision over a time horizon, defined as: 
\begin{align}
    \mathcal{C}(\bm\Xi_{\Psi})=&\frac{1}{T}\int\displaylimits_{0}^{T}\mathcal{C}(\bm{\Xi}_{\Psi}(t))\mathrm{d}t\\=&\frac{1}{T}\int\displaylimits_{0}^{T}\int\displaylimits_{\mathbb{R}^2}p(m=1|\vec{x})p(\vec{x}|\Psi,t)\mathrm{d}\vec{x}\mathrm{d}t.\label{Costeqn}
\end{align}
The probability of collision at some time and location is computed by taking the product of the probability of the trajectory reaching the coordinate at the given time and the probability that the coordinate is occupied. We are provided with an occupancy map and the probability a coordinate of interest, $\vec{x}\in \mathbb{R}^{2}$, being occupied is denoted as $p(m=1|\vec{x})$. To evaluate the cost, we could marginalise over the all trajectory parameters, but this may be difficult due to the high dimensionality of $\bm{\omega}$, instead work in the 2-D world space, where the environmental constraints are defined. As $\bm\Xi_{\Psi}$ is constructed by taking the dot product of weight vectors and feature vectors, trajectory coordinate distribution in world space at time $t$ is:
\begin{align}
p(\bm{\Xi}_{\Psi}(t))&\!=\!\sum_{r=1}^{R}\!\alpha_{r}\overbrace{{\mathrm{MN}(\mathcal{M}_{r}^{\top}\bm{\phi}(t),\mathcal{V}_{r},\bm{\phi}(t)^{\top}\mathcal{U}_{r}\bm{\phi}(t)\!)}}^{\text{matrix normal distribution}}\\&\!=\!\sum_{r=1}^{R}\!\alpha_{r}\overbrace{\mathcal{N}(\underbrace{\mathcal{M}_{r}^{\top}\bm{\phi}(t)}_{=:\bm{\mu}_{r}},\underbrace{\bm{\phi}(t)^{\top}\mathcal{U}_{r}\bm{\phi}(t)\mathcal{V}_{r}}_{=:\Sigma_{r}}).}^{\text{multivariate normal}}\label{worldspacedist}
\end{align}
By the equivalences between matrix normal and multivariate normal distributions \cite{Dawid1981SomeMD}, we have a mixture of two-dimensional multivariate normal distributions in world-space, $\bm{\Xi}_{\Psi}(t)\sim\sum_{r=1}^{R}\alpha_r\mathcal{N}(\bm{\mu}_r,\Sigma_r)$. This is the probability distribution over trajectory coordinates at time $t$, $p(\vec{x}|\Psi,t)$. The mixture means and covariances of each component are defined as 
$
{\bm{\mu}_r:=\mathcal{M}_{r}^{\top}\bm{\phi}(t)},$ and $ \Sigma_{r}:={\bm{\phi}(t)^{\top}\mathcal{U}_{r}\bm{\phi}(t)\mathcal{V}_{r}}.
$
Substituting \cref{worldspacedist} into the inner integral of \cref{Costeqn}, gives the cost operator equation at given time:
\begin{align}
&\mathcal{C}(\bm{\Xi}_{\Psi}(t))=\sum_{r=1}^{R}\alpha_{r}\nonumber\\ &\ldots\int\displaylimits_{\mathbb{R}^{2}}\Big\{\frac{\exp[-\frac{1}{2}(\vec{x}-\bm{\bm{\mu}_r})^{\top}\Sigma_{r}^{-1}\!(\vec{x}-\bm{\mu}_r)]}{(2\pi)|\Sigma_{r}|^{\frac{1}{2}}}\Big\}\underbrace{p(m=1|\vec{x})}_{\text{From map}}\mathrm{d}\vec{x}\label{costEqnTime}
\end{align}
Note that $\bm{\mu}_r$ and $\Sigma_{r}$ are dependent on $\mathcal{M}_r,\mathcal{U}_r,\mathcal{V}_r$. During the optimisation, we assume that each mode can be optimised independently and that the importance of each mode, represented by $\alpha$, is accurately captured by the learning model and does not change during trajectory optimisation. We then define the set of parameters to optimise as $\Psi=\big\{\mathcal{M}_{r},\mathcal{U}_{r},\mathcal{V}_{r}\}_{r=1}^{R}$.
%\begin{figure}[h]
%\floatbox[{\capbeside\thisfloatsetup{capbesideposition={right,top},capbesidewidth=0.3\textwidth}}]{figure}
%{\caption{The abscissae (black) for estimating the distribution in world space with the Gauss-Hermite quadrature scheme}\label{fig:absc}}
%{\includegraphics[width=0.42\textwidth]{xydist.png}}
%\end{figure}
The integral defined in \cref{costEqnTime} may be intractable to evaluate analytically, so we aim to find an approximation to it. The coordinate distribution at a given time in world space, $\bm{\Xi}_{\Psi}(t)$, is a mixture of bi-variate Gaussian distributions. We use Gauss-Hermite quadrature, suitable for numerically integrating exponential-class functions, to approximate \cref{costEqnTime}. Quadrature methods approximate an integral by taking the weighted sum of integrand values sampled at computed points, called abscissae. We introduce the change of variables, $\bm{z_{r}}=\frac{1}{\sqrt{2}}L_{r}^{-1}(\vec{x}-\bm{\mu}_r)$, where the Cholesky factorisation gives $\Sigma_{r}=L_{r}L_{r}^{T}$. Using Gauss-Hermite quadrature schemes, we have:
\begin{align}
    \mathcal{C}(\bm\Xi_{\Psi}(t))
    %=\sum_{r=1}^{R}\!\alpha_r\!\int\displaylimits_{\mathbb{R}^{2}}\!\frac{\!p(\!m\!=\!1|\!\sqrt{2}L\bm{z_{r}}\!+\!\bm{\mu}_r)\!\exp(\!-\bm{z_{r}}^{\top}\!\bm{z_{r}}\!)|\!\sqrt{2}L\!|}{\sqrt{2\pi\Sigma_r}}\mathrm{d}\vec{x}\nonumber\\
    %&=\sum_{r=1}^{R}\!\alpha_r\!\int\displaylimits_{\mathbb{R}^{2}}\pi^{-1}\!p(m\!=\!1|\!\sqrt{2}\!L\bm{z_{r}}\!+\!\bm{\mu}_r)\!\exp(\!-\bm{z_{r}}^{\top}\!\bm{z_{r}}\!)\mathrm{d}\vec{x}\nonumber
    &\approx\sum_{r=1}^{R}\alpha_r\pi^{-1}\sum_{i=1}^{I}\sum_{j=1}^{J}\beta^{i}\beta^{j}p(m=1|\sqrt{2}L\bm{z}+\bm{\mu}_r),\label{eqnQuad}
\end{align}
where we find the values of $\bm{z}^{i,j}=[z^{i},z^{j}]^{\top}$ as abscissa obtained from roots of Hermite polynomials, and $\beta^{i}$, $\beta^{j}$ are the associated quadrature weights. Details of abscissae and weight calculations, as well as a review on multi-dimensional Gauss-Hermite quadrature, can be found in \cite{GaussHermite}. An example of abscissae points when estimating a multi-variate Gaussian is shown in \cref{fig:absc}. Next, we evaluate the outer integral in \cref{Costeqn} using rectangular numerical integration for ${\mathcal{C}(\bm\Xi_{\Psi})=\int_{0}^{T}\mathcal{C}(\bm{\Xi}_{\Psi}(t))}\mathrm{d}t$. With $\mathcal{C}(\bm\Xi_{\Psi})$ defined, we are in a position to solve the combined learning and constrained optimisation problem, defined earlier in \cref{formulationWhole}, using a non-linear optimisation solver, such as the sequential least squares optimiser SLSQP \cite{slsqp}. Analytical gradients with respect to occupancy is provided when using a continuous differential occupancy map such as those introduced in \cite{HilbertMap, icra_wzhi2019}. By utilising the equivalence between matrix normal distributions and multivariate Gaussians \cite{MLEMatN}, the KL divergence term between $\bm{\omega}_{\Psi}$ and $\bm{\omega}_{\Psi_{0}}$ components can be computed \cite{BishopBook}. After obtaining the solution of the optimisation we have the optimised distribution of $\bm{\omega}_{\Psi}^{*}$. By taking dot products between $\bm{\omega}_{\Psi}^{*}$ and $\bm{\phi}$, we can construct $\bm{\Xi}_{\Psi}^{*}$ the constraint-compliant trajectory distribution of futures. Individual realised trajectories $\bm{\xi}$ can then be generated from $\bm{\Xi}_{\Psi}$. 

Although we have focused on collision constraints with respect to environment structure, the optimisation allows for constraints on velocity or acceleration. The continuous nature of our trajectories allows time derivatives of trajectories to be obtained. Specifically, the \emph{derivative reproducing property}, ensures the derivatives of smooth kernels also correspond to kernels \cite{DerivativeRKHS}. We can reason about the $n^{th}$-order derivatives of displacement by replacing $\bm{\phi}(t)$ with $\frac{\partial^{n}\bm{\phi}(t)}{\partial t^{n}}$. 

%\begin{align}
%    \quad\mathrm{D}_{KL}(\bm{\omega}^{r}_{\Psi}\lVert\bm{\omega}^{r}_{0})=\frac{1}{2}\{\log\frac{|\mathcal{V}^{r}_{0}\otimes\mathcal{U}^{r}_{0}|}{|\mathcal{V}^{r}_{\Psi}\otimes\mathcal{U}^{r}_{\Psi}|}-2M&+\mathrm{tr}[(\mathcal{V}^{r}_{0}\otimes\mathcal{U}^{r}_{0})^{-1}\mathcal{V}^{r}_{\Psi}\otimes\mathcal{U}^{r}_{\Psi}]\nonumber\\
%    &+(\mathcal{M}^{r}_{0}-\mathcal{M}^{r}_{\Psi})^{\top}(\mathcal{V}^{r}_{0}\otimes\mathcal{U}^{r}_{0})^{-1}(\mathcal{M}^{r}_{0}-\mathcal{M}^{r}_{\Psi})\}.
%\end{align}
%The architecture of our neural network is: $\mathrm{Dense}(15\times R \times M)\rightarrow \mathrm{Dense}(5\times R \times M) \rightarrow \mathrm{Dense}(5\times R \times M) $.

\section{Experimental Evaluation}
We empirically evaluate (1) the ability of our learning component to learn probabilistic representations of future trajectories to provide good prior trajectories distributions; (2) the ability of the proposed framework to enforce collision constraints, and its effect on trajectory quality.

\subsection{Datasets and Metrics}
We run our experiments on one simulated dataset, and two real-world datasets, including:
\begin{itemize}[leftmargin=*]
    \item Simulated dataset \textbf{(Simulated)}: This contains simulated trajectories on a floor-plan;
    \item TH{\"O}R dataset \cite{thorDataset2019} \textbf{(Indoor)}: This dataset contains real human trajectories walking in a room with three large obstacles in the center of the room. 
    \item Lankershim dataset \textbf{(Traffic)}: This dataset contains real vehicle data along a long boulevard. We use a particular busy intersection between $x=[-80, 80]$ and $y=[250, 500]$. 
\end{itemize}
Throughout each of our experiments, we ensure that the observed trajectory conditioned on contains 10 timesteps. The prediction horizons, $T$, for the simulated, indoor and traffic datasets are 15, 10, and 20 timesteps respectively. We split long trajectories, such that we have 200 pairs of partial trajectories in the simulated dataset, while the indoor dataset contained 871 pairs of partial trajectories, and the traffic dataset contained 6175 pairs. Trajectories that were shorter than the length of the prediction horizon were filtered out. A train-to-test ratio of 80:20 was used for each dataset. For our learning model, we set the number of mixture components $R=2$, hyper-parameters $M=8, 10, 11$, and $\gamma=0.05, 0.1, 0.2$ for the simulated, indoors, and traffic datasets respectively. Maps for the simulated and indoor datasets are available, and we construct a map for the traffic dataset by considering the road positions. Our occupancy map is represented as a continuous Hilbert Map \cite{HilbertMap}, giving smooth occupancy probabilities over coordinates, and occupancy gradients. 

Metrics used in the quantitative evaluation include: (i) \textbf{Average displacement error (ADE)}: The average euclidean distance error between the expectation of the nearest distribution component of trajectories with ground truth trajectory; (iii)  \textbf{Final displacement error (FDE)}: The euclidean distance error at the end of time horizon between the expectation of the nearest distribution component of trajectories with ground truth trajectory. (iii)  \textbf{Average Likelihood (AL)}: To take into account the uncertain nature of motion predict, for probabilistic models, we record the likelihood of drawing the trajectory averaged over the timesteps. Note that likelihoods are difficult to interpret and not comparable over different datasets. Lower ADE and FDE, and higher AL indicate better performance. For the evaluation of collisions, we also record the percentage of constraint-violating trajectory distributions in test set.

\subsection{Learning Continuous-Time Trajectory Distributions as Priors}
The trajectory distribution learning component of our framework provides a prior distribution for the optimisation. We examine whether the learning component of our framework is able to learn distributions of future trajectories from data, comparing the performance with several benchmark models. The benchmark models are: (1) Kernel trajectory maps (KTMs) \cite{wzhiKTM}, (2) Constant velocity (CV) model, and (3) Naive neural-network (NN) model, where we learn a mapping between past trajectories and future trajectories by minimising the mean squared error loss. The neural network comprised of Relu-activated fully-connect layers with number of neurons: $(560)\rightarrow(180)\rightarrow(180)\rightarrow(N_t)$, where $N_t$ gives the trajectory time horizon. 

\begin{figure}[]
\centering
        \includegraphics[width=0.4\textwidth]{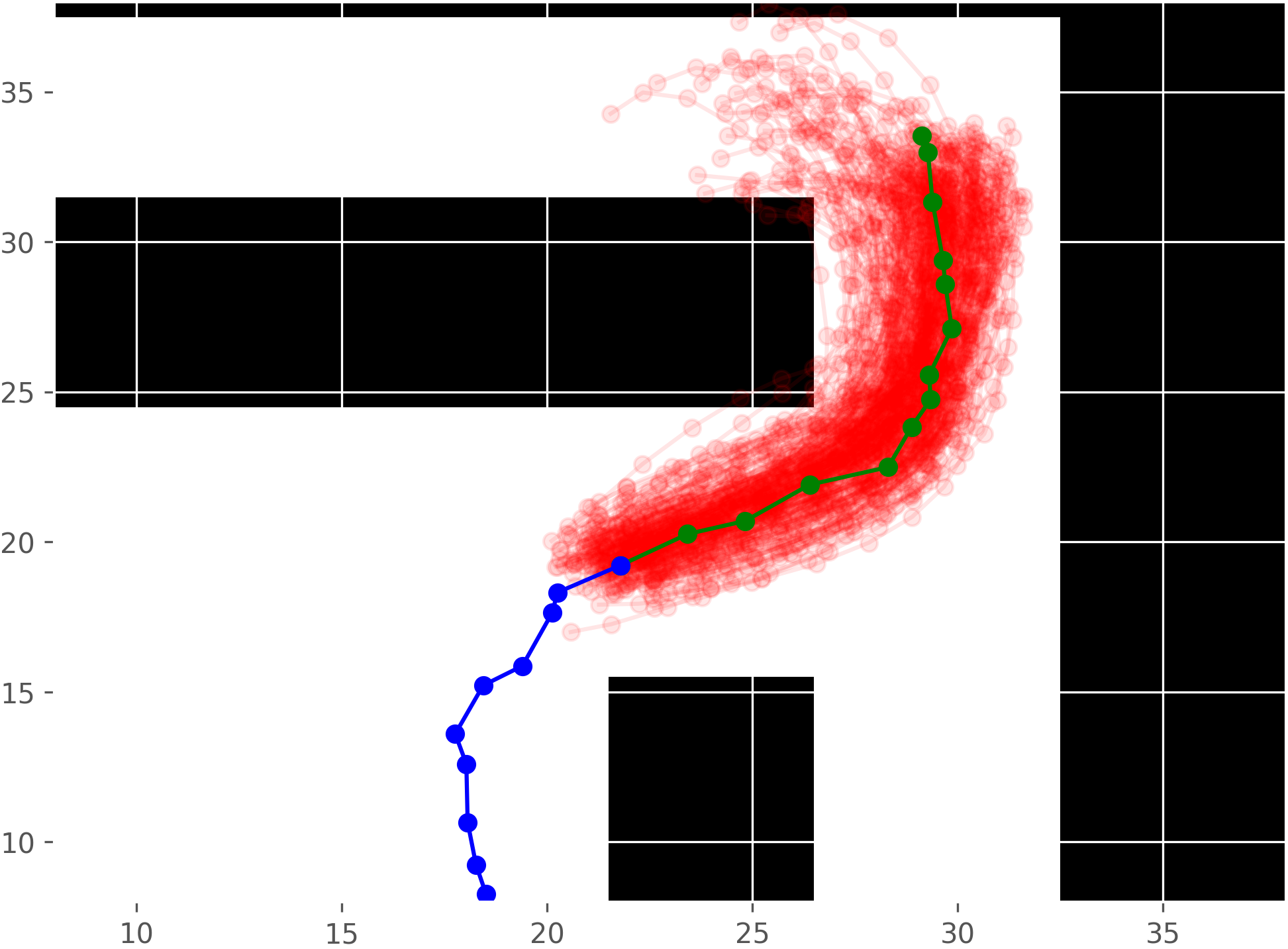}
    \hspace{5mm}
        \includegraphics[width=0.4\textwidth]{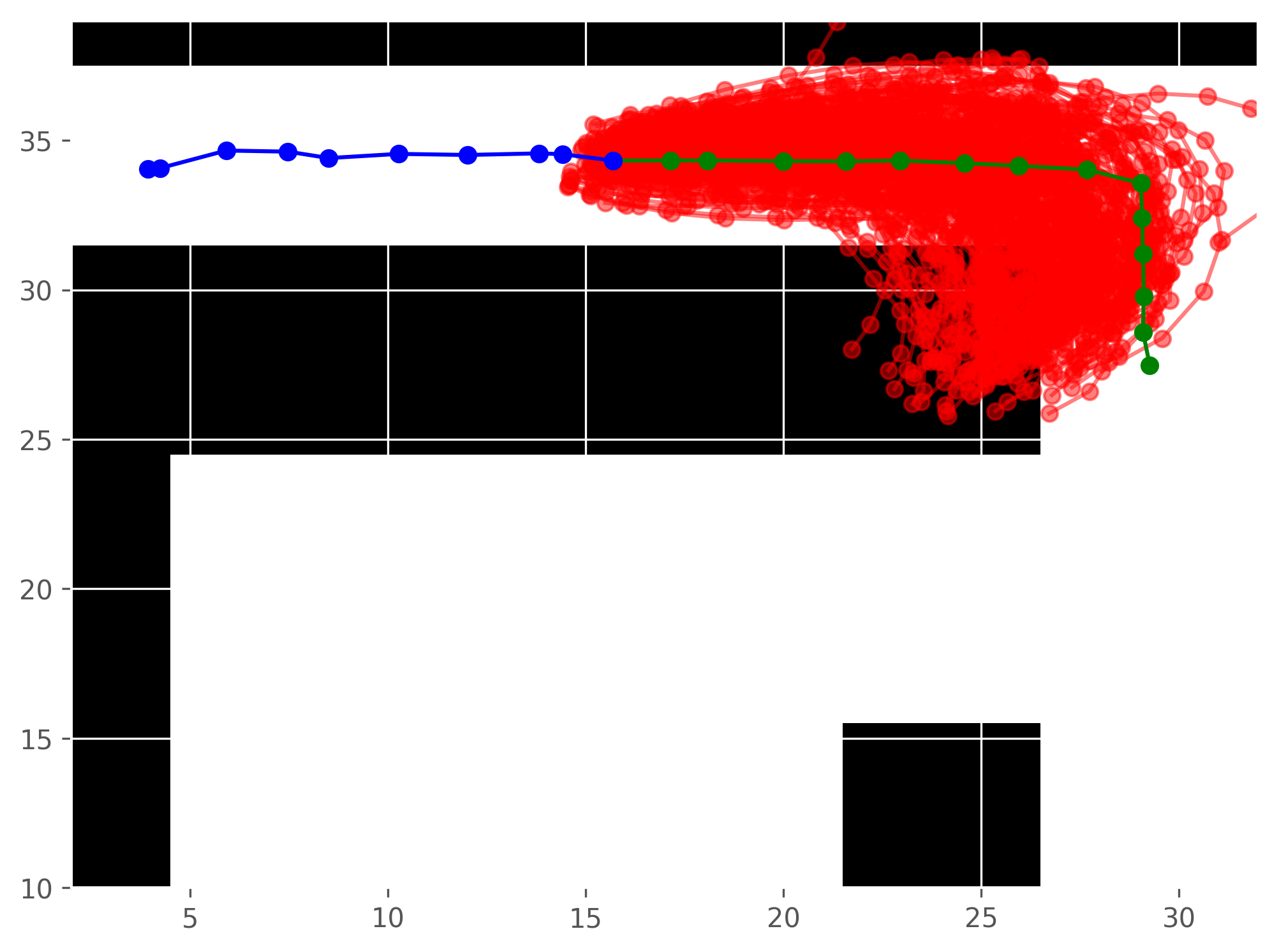}

        \includegraphics[width=0.4\textwidth]{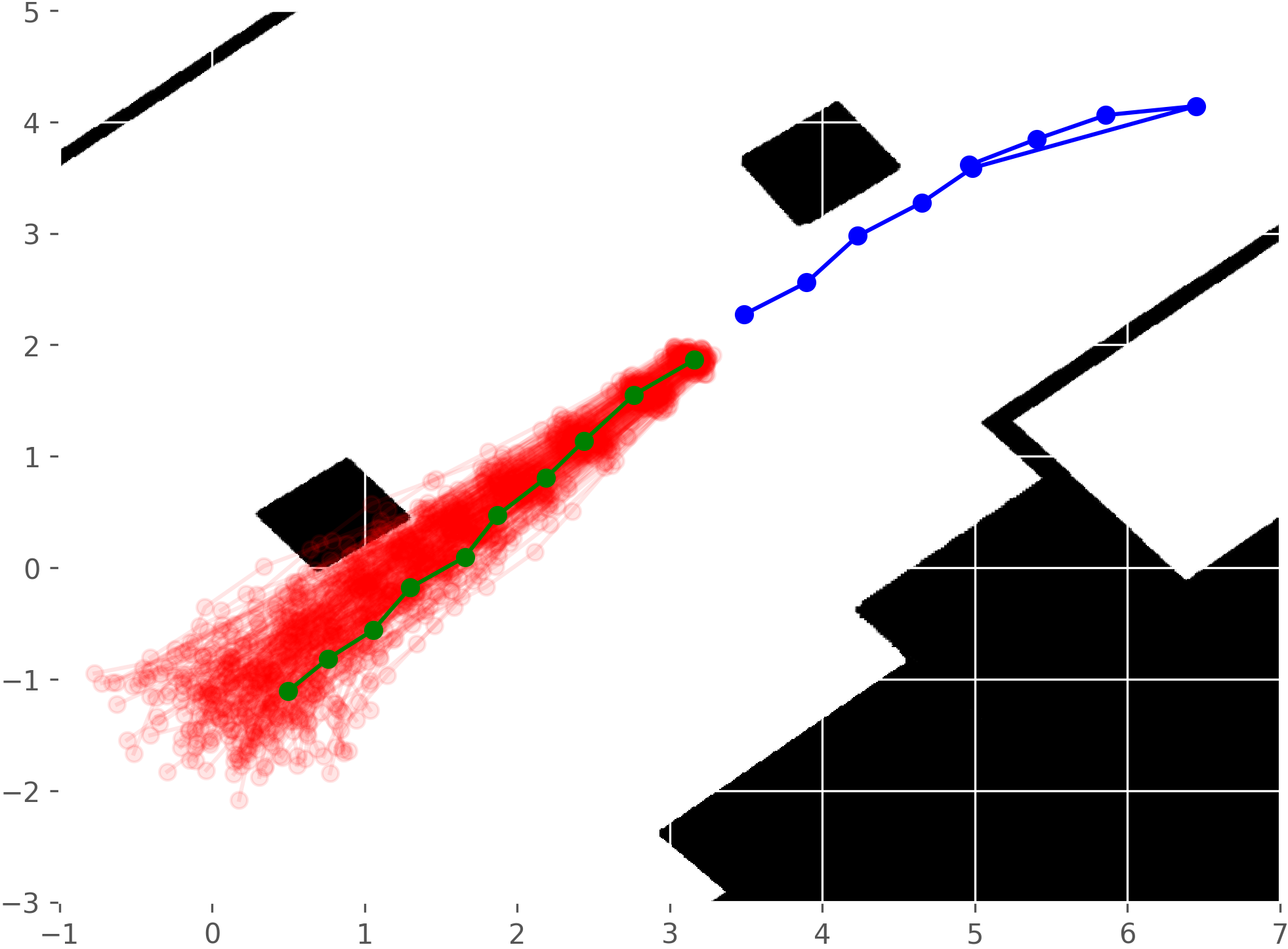}
    \hspace{5mm}
        \includegraphics[width=0.4\textwidth]{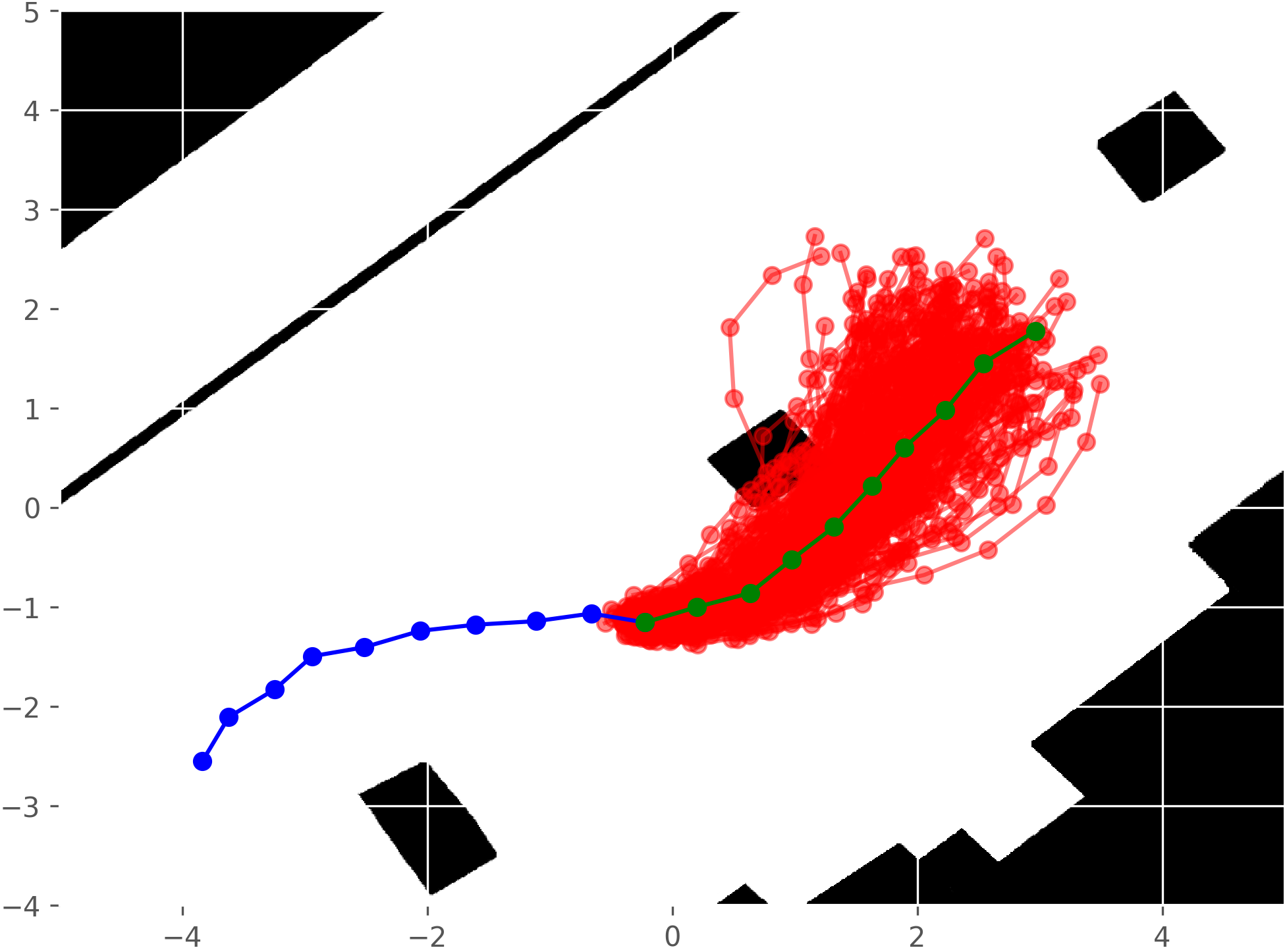}

        \includegraphics[width=0.4\textwidth]{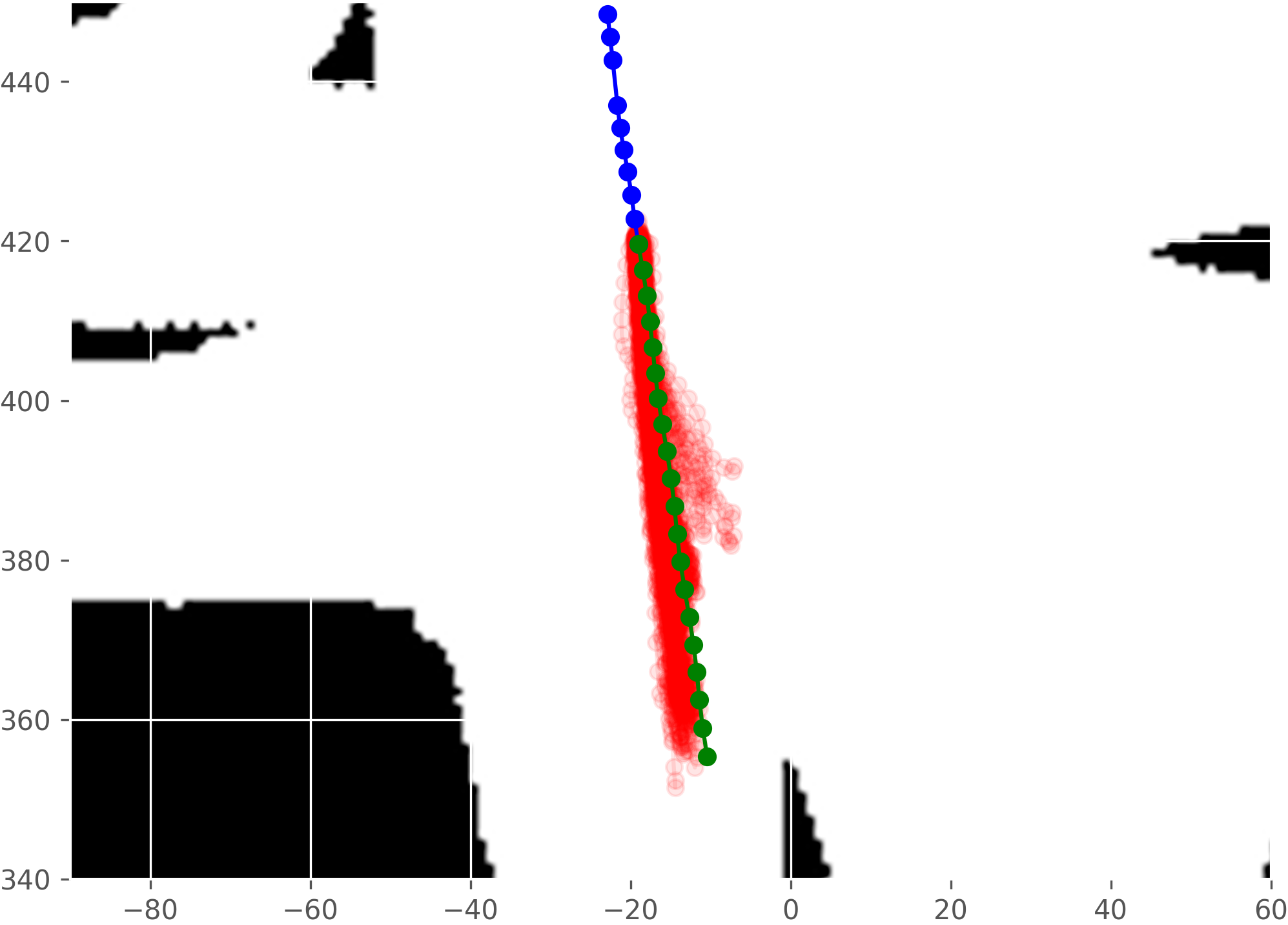}
    \hspace{5mm}
        \includegraphics[width=0.4\textwidth]{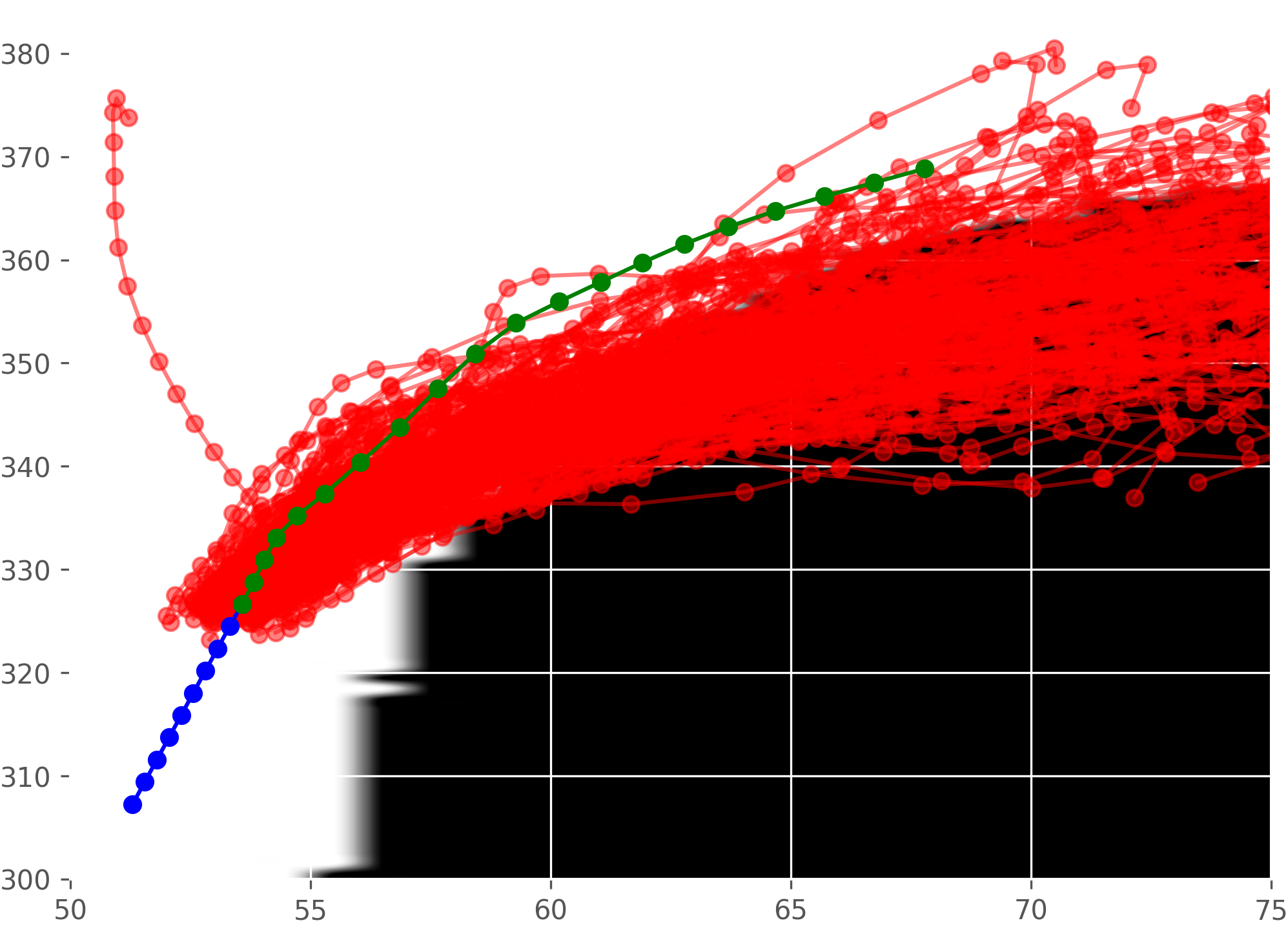}
        
\captionof{figure}{Quantitative evaluations of learned distributions of trajectories (red), Conditioned observations (blue), ground truth (green). We see that our learning model gives relatively good priors, but many sampled trajectories collide with obstacles. The prior distributions are further enforced to be collision-free (See \cref{xyTraj})}\label{ExamplePred}
\end{figure}

We see that the learning component of our framework is able to learn high quality distributions of trajectories. \Cref{ExamplePred} shows examples of extrapolated trajectory distributions in red, and the observed trajectory in blue. We see that the learned distribution is relatively close to the ground truth green trajectory, and is able to capture the uncertain nature of motion prediction. Although a non-negligible number of sampled trajectories collide with the environment. \Cref{tableRes1} summarises the performance results of our method against our benchmarks. We see that our learning model is comparable to the method proposed in \cite{wzhiKTM}, while outperforming a constant velocity and naive neural network model. While the KTM method performs slightly better on the traffic dataset, it is purely learning-based and lacks mechanisms to enforce constraints. We note that our method encodes the observed trajectory by simply flattening the input coordinates, as described in \cref{Learndists}. More sophisticated methods of encoding the observed trajectory (such as with LSTM networks \cite{LSTM}), independent of our outlined output of trajectory distributions, could improve the performance of the learning model.

%\begin{adjustbox}{width=0.9\linewidth}
%While the KTM method performs slightly better on the traffic dataset, it is not compatible with the constraint optimisation aspect of this work and as such is not a suitable basis.
\begin{table}
\centering
\begin{tabular}{llccc}
\toprule
                           &          & \small ADE  & \small FDE   & \small AL \\
                           \midrule
\multirow{4}{*}{\small Simulated} &\small Ours      & 1.29 & 2.27  & 0.12       \\
                           &\small KTM \cite{wzhiKTM}     & 1.44 & 2.41  & 0.11       \\
                           &\small CV       & 4.27 & 11.66 & -          \\
                           &\small NN-naive & 2.11 & 6.07  & -          \\
                           \midrule
\multirow{4}{*}{\small Indoors}   &\small Ours      & 0.94 & 1.32  & 3.32       \\
                           &\small KTM      & 0.65 & 1.26  & 2.88       \\
                           &\small CV       & 2.08 & 4.66  & -          \\
                           &\small NN-naive & 1.84 & 3.35  & -          \\
                           \midrule
\multirow{4}{*}{\small Traffic}   &\small Ours      & 4.99 & 7.24  & 0.04       \\
                           &\small KTM      & 4.98 & 7.03  & 0.043      \\
                           &\small CV       & 5.14 & 10.38 & -          \\
                           &\small NN-naive & 21.1 & 39.5  & -      \\ 
                           \bottomrule
\end{tabular}\captionof{table}{We evaluate the quality of our learned trajectory prior with benchmark models. Lower ADE and FDE, and higher AL indicate better performance.}\label{tableRes1} 
\end{table}

\begin{figure*}[t]
        \begin{subfigure}[]{0.3\textwidth}
        \centering
        \includegraphics[width=\textwidth]{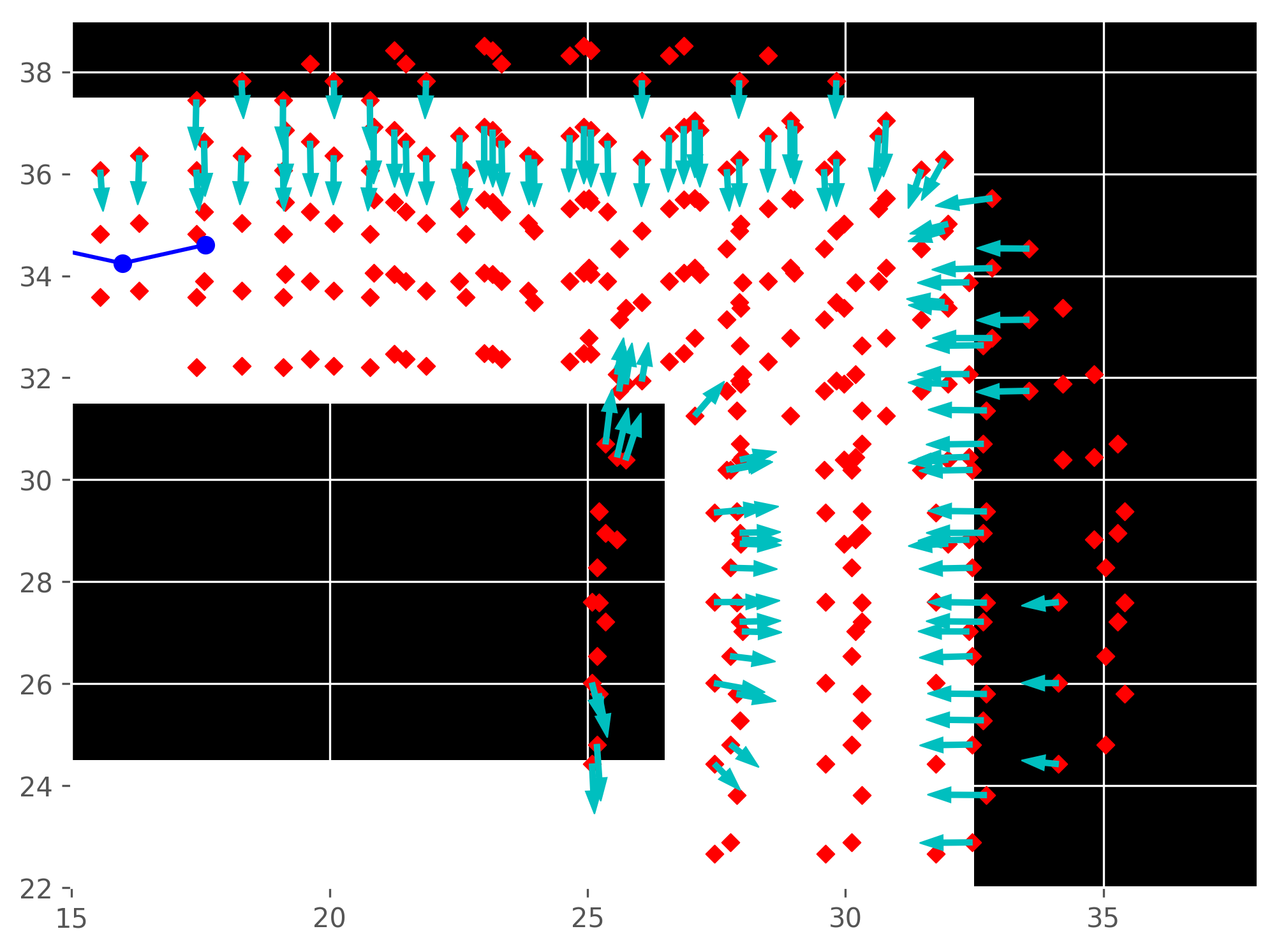}
    \end{subfigure}%
    \hspace{2mm}
    \begin{subfigure}[]{0.3\textwidth}
        \centering
        \includegraphics[width=\textwidth]{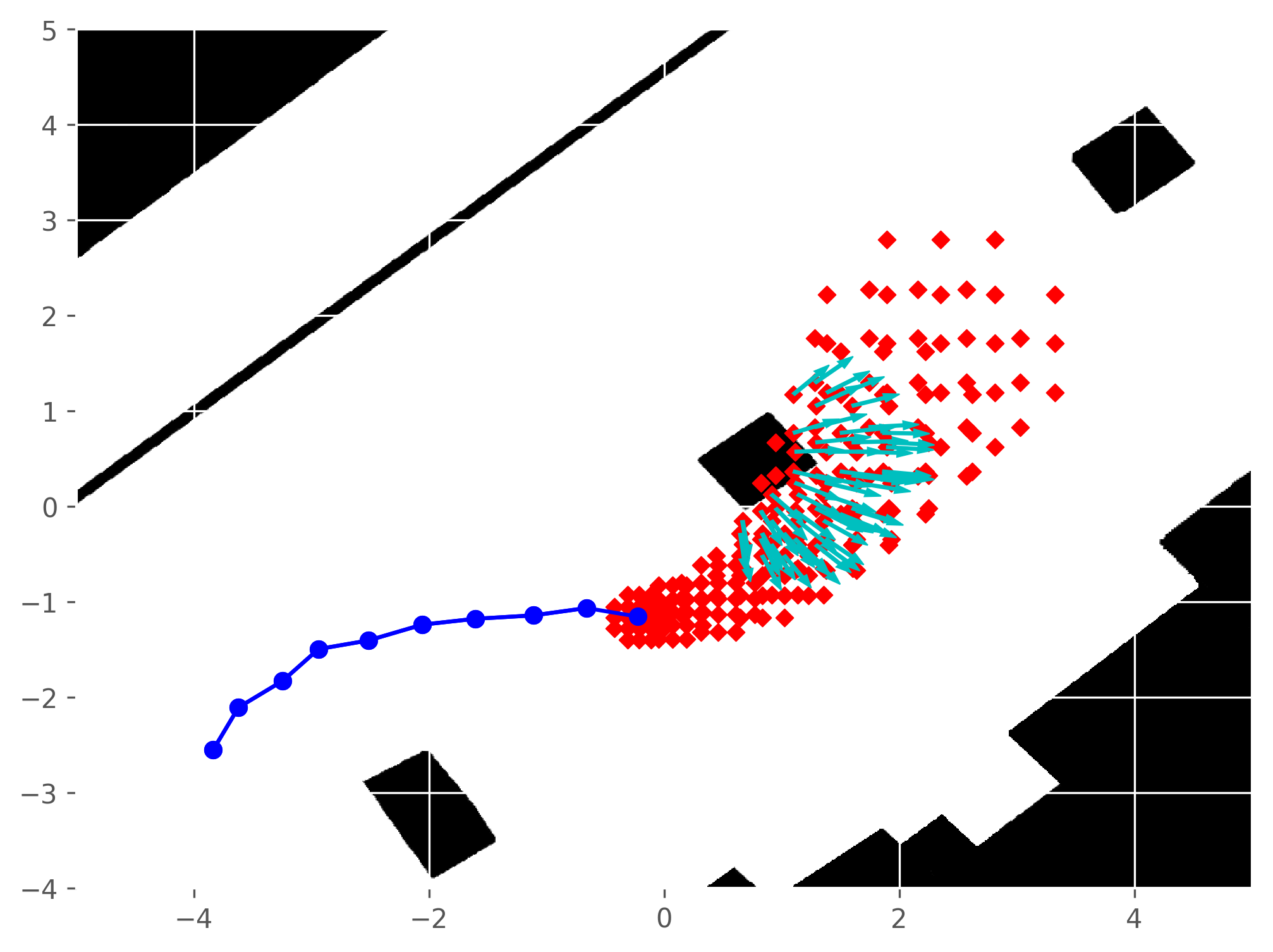}
    \end{subfigure}%
    \hspace{2mm}
    \begin{subfigure}[]{0.3\textwidth}
        \centering
        \includegraphics[width=\textwidth]{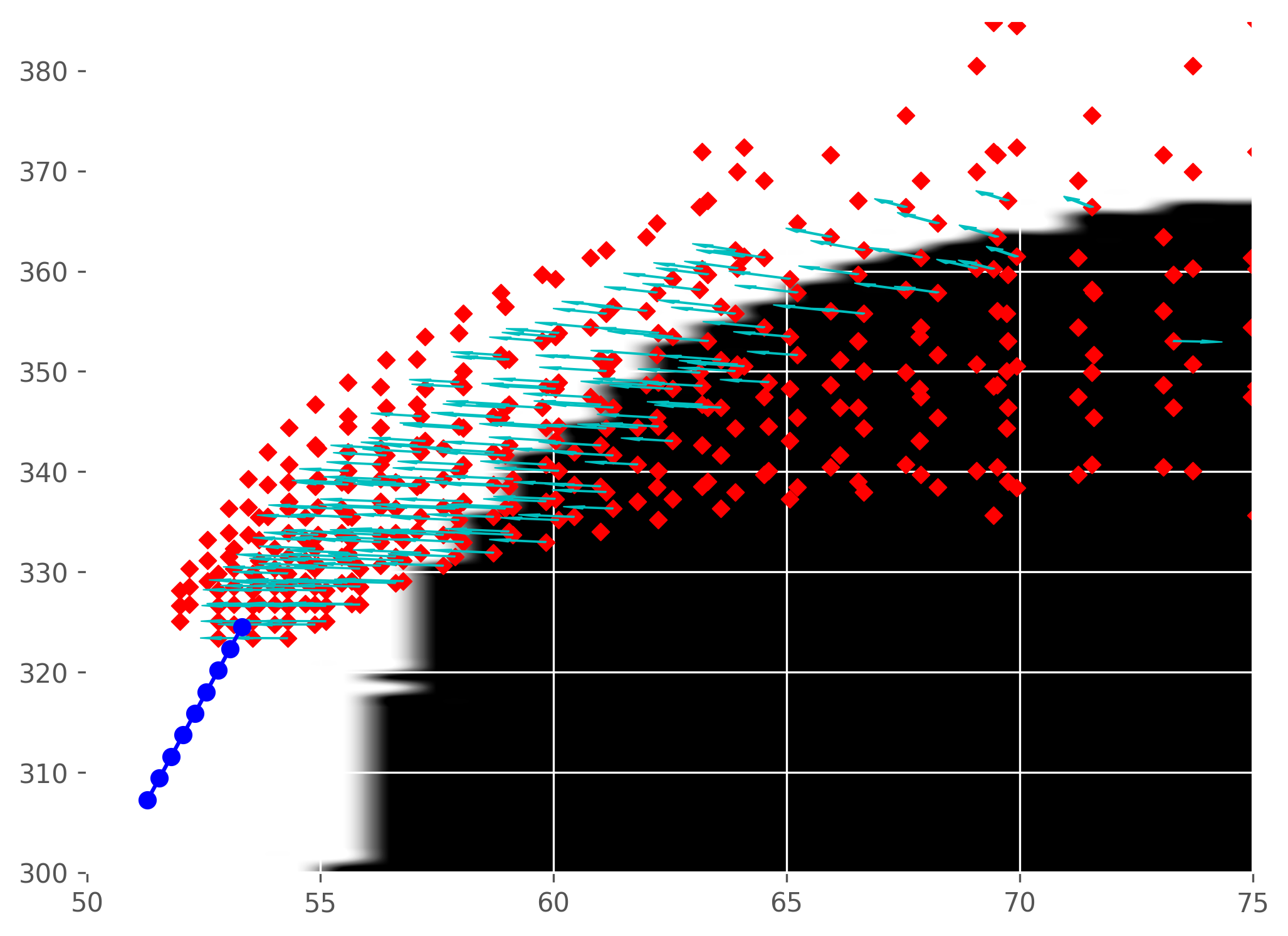}
    \end{subfigure}%
    
     \begin{subfigure}[]{0.3\textwidth}
        \centering
        \includegraphics[width=\textwidth]{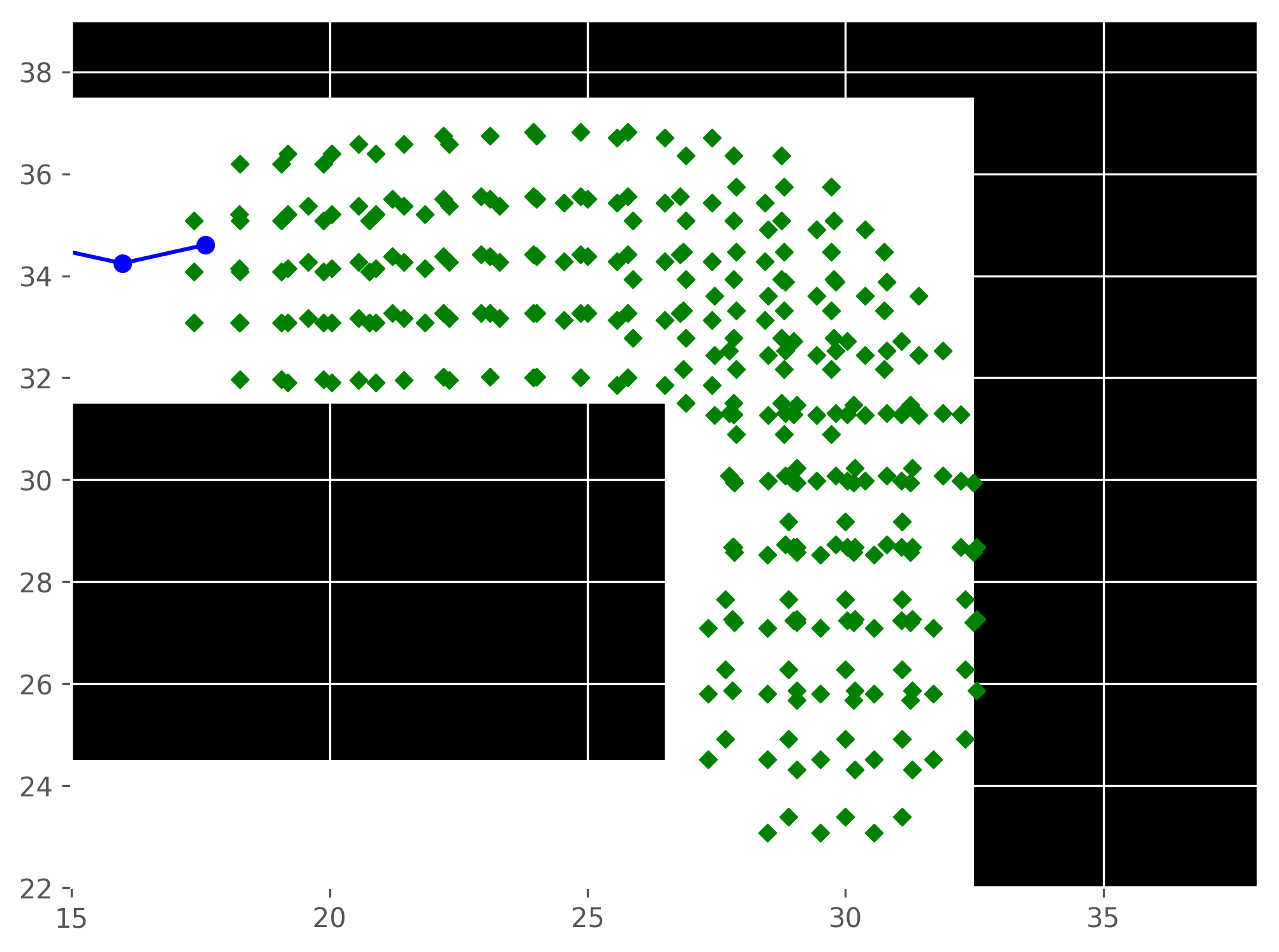}
    \end{subfigure}%
    \hspace{2mm}
     \begin{subfigure}[]{0.3\textwidth}
        \centering
        \includegraphics[width=\textwidth]{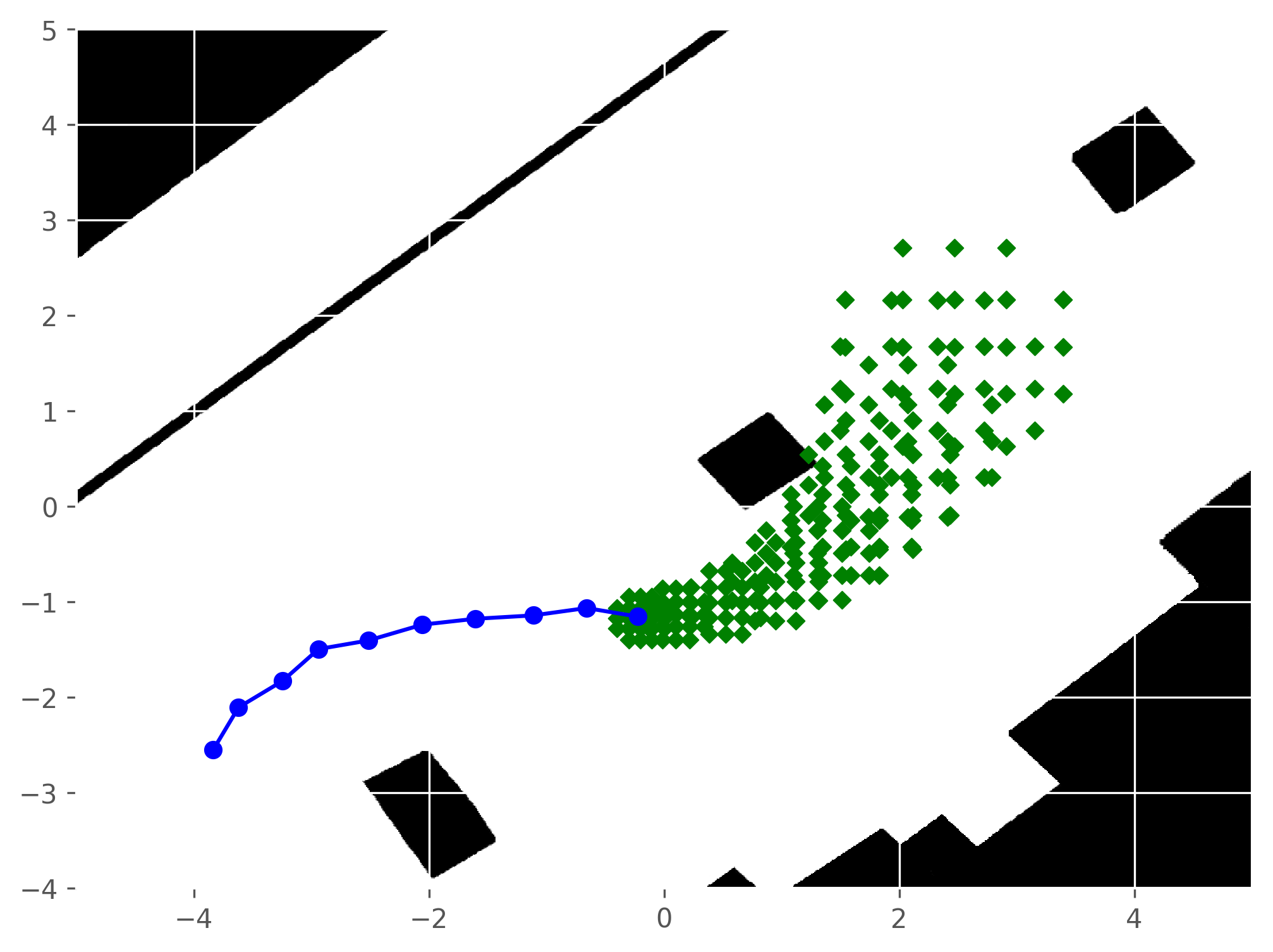}
    \end{subfigure}%
    \hspace{2mm}
     \begin{subfigure}[]{0.3\textwidth}
        \centering
        \includegraphics[width=\textwidth]{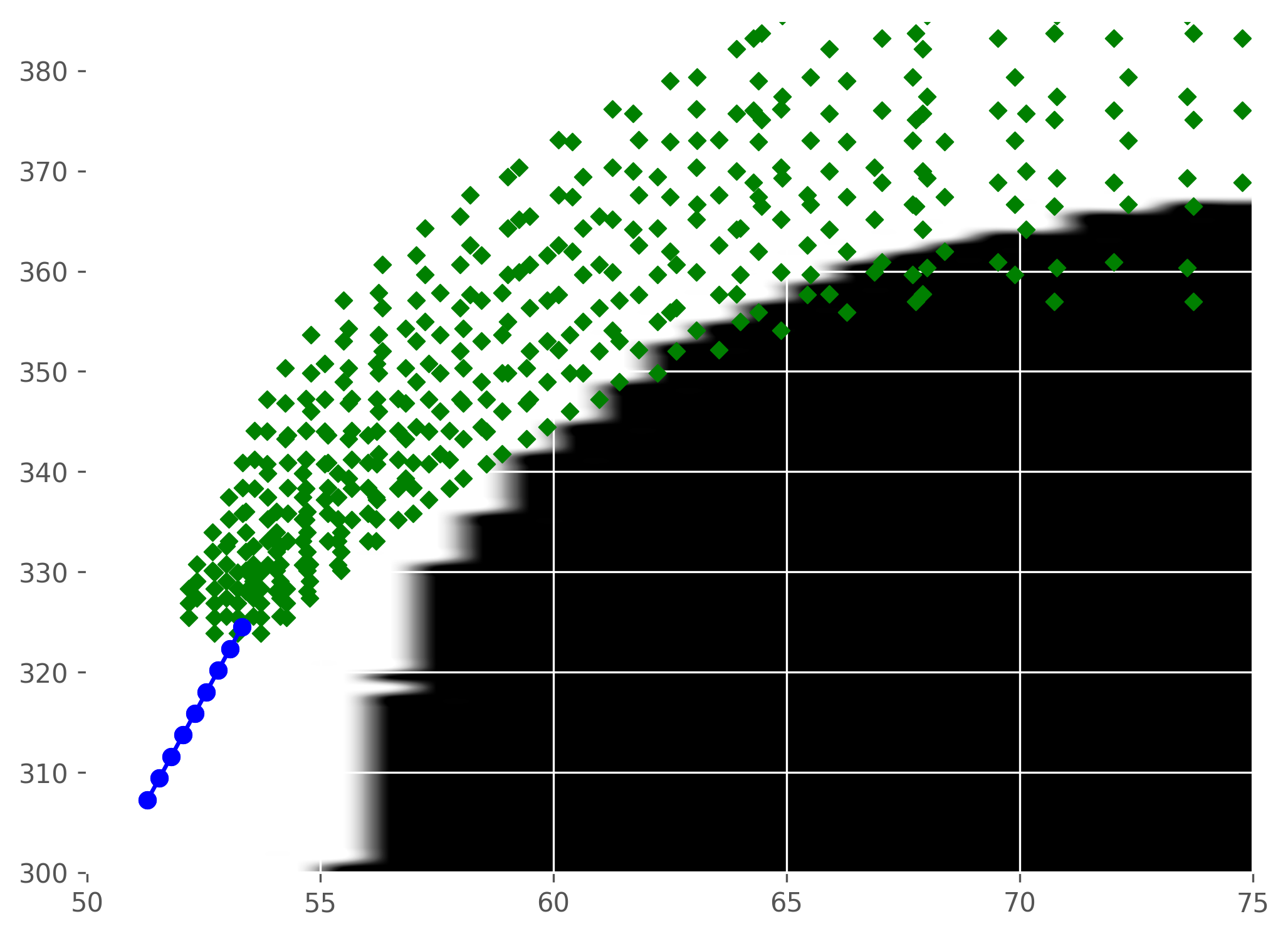}
    \end{subfigure}%
    \caption{Examples from each dataset of the optimisation component of the framework applying constraints. The top row shows abscissae points of the main component, with notable gradients with respect to occupancy shown as arrows (teal) at the abscissae. The bottom row shows the same abscissae points after optimisation. We see that the constrained distribution subtly conforms better to structures in the environment, while still closely resembling the learned trajectory distribution. The left, middle and right sub-figures are examples from the simulated, indoors, and traffic datasets respectively.}\label{xyTraj}

\end{figure*}

\subsection{Enforcing Occupancy Compliance via Trajectory Optimisation}

After obtaining a prior from our learned model, we apply constraints on the prediction via trajectory optimisation. We will evaluate whether collision constraints can be enforced, and their effect on trajectory quality. We define the constraint such that the time-averaged probability of collision to be less than or equal to 0.05, $\mathcal{C}-0.05\leq 0$. The SLSQP optimiser was able to solve the optimisation quickly, under half a second, with a python implementation. We obtain trajectory distribution priors from our learned model, and select all of the observations where the learned prior trajectory is constraint-violating. We evaluate the performance of the trajectory priors, and compare them with the trajectory distribution predictions after enforcing constraints. The quantitative results are listed in \cref{tableRes2}. We see after solving with structural constraints, all predictions are feasible. Additionally, the trajectory distribution quality improves after enforcing constraints for all the datasets considered.

\begin{table}[]
\small
\centering
\begin{tabular}{llcccc}
\toprule
          &             & \small ADE  & \small FDE   & \small AL    & \small C.V.P\\
          \midrule
\small Simulated & \small Unoptimised & 1.48 & 2.87  & 0.11  & 40\%                 \\
          & \small Optimised   & 1.28 & 2.33  & 0.17  & 0\%                  \\
\small Indoors   & \small Unoptimised & 1.62 & 2.41  & 3.42  & 15.4\%               \\
          & \small Optimised   & 1.44 & 2.38  & 3.3   & 0\%                  \\
\small Traffic   & \small Unoptimised & 6.56 & 11.81 & 0.034 & 5.7\%                \\
          & \small Optimised   & 5.14 & 8.95  & 0.043 & 0\%\\
          \bottomrule
\end{tabular}
\caption{We evaluate the quality of our optimised trajectory distribution with collision constraints, relative to the prior. We see that providing constraints improves prediction quality across all dataset. The percentage of trajectory distributions violating constraint, $\mathcal{C}-0.05\leq 0$, relative to the entire test set is also given, as Constraint Violation Percentage (C.V.P). After optimisation all trajectory distributions are constraint-compliant.}\label{tableRes2}
\end{table}

Qualitative results are shown in \cref{xyTraj}. We see the abscissae points of the largest trajectory distribution component before (top) and after (bottom) the trajectory optimisation of examples in each dataset. Our framework is able to ensure that the distributions comply with environmental structure. Notably, we see that the trajectory optimisation not only adjusts the mean of the trajectory distribution, but also optimises the variances, giving more robust trajectory distributions. This effect is most prominent in the left sub-figure, where the predicted distribution of trajectories in a simulated hallway recovers a much tighter variance which follows the environment structure. We also observe instances collision-avoidance with objects in open space (demonstrated in middle sub-figure), as well as increased adherence to road structure (right sub-figure).

%We now investigate whether our optimisation component is able to use the learned trajectory distribution, and an occupancy representation, to ensure our trajectories are consistent with the environment structure. We run our experiments on simulated data containing an indoor occupancy map and associated trajectories presented in \cite{occtraj120}. Upon training the learning component, we optimise and constrain the average probability of over time of being in a occupied region to be less than or equal to $1\%$, i.e. we set $\epsilon=0.01$ for \cref{formulationWhole}. 

%A visual demonstration of combined learning and optimisation is shown in \cref{trajOpt}. As the neural network learning component aims to mimic training data and does not explicitly enforce avoidance with occupied regions, the initial learned distribution of trajectories has a relatively high probably of sampling collision-prone trajectories. After optimising the distribution, we get a relatively collision-free distribution of trajectories. We also note that the learned distribution is not far off from the optimised distribution, and is used to ``warm-start'' the optimisation.
  
\section{Conclusion}
We introduced a novel framework to learn motion predictions from observations while conforming to constraints, such as obstacles. Our framework consists of a learning component which learns a distribution over future trajectories. This distribution is used as a prior in a trajectory optimisation step which enforces chance constraints on the trajectory distribution. We empirically demonstrate that our framework can learn complex trajectory distributions and enforce compliance with environment structures via optimisation. This results in reduced variance and the avoidance of obstacles by the predicted trajectories, leading to improved prediction quality.
% We introduced a novel framework to obtain distribution of trajectories that comply with constraints, in particular environment structure constraints, for trajectory prediction problems. Our framework consists of a learning component to obtain at a prior distribution of trajectories, and a trajectory optimisation component to impose constraints on a distribution of trajectories. We empirically demonstrate that our framework can learn complex probabilistic trajectory distributions, and enforce compliance with the environment structures via optimisation. We observe behaviours such as reduced variance in hallway structures and avoidance of obstacles in predicted trajectory distributions, resulting in improved prediction quality on both real and simulated datasets.  

\bibliographystyle{ieeetr}
\bibliography{references}

\end{document}